\documentclass[10pt,twocolumn,letterpaper]{article}

\usepackage{iccv}              %

\usepackage{epsfig}
\usepackage{graphicx}
\usepackage{amsmath}
\usepackage{amssymb}
\usepackage{tensor}
\usepackage{tikz}
\usepackage{mathtools}
\usepackage{threeparttable}
\usepackage{siunitx} 
\usepackage{booktabs}
\usepackage{multirow}
\usepackage{placeins}
\usepackage{pgffor}
\usepackage{pgfplots}
\usepackage{tikzscale}
\usepackage{standalone}
\usepackage{bm}
\usepackage{colortbl}
\usepackage{pifont}
\usepackage{tabularray}
\usepackage{textcomp}
\usepackage{gensymb}
\usepackage{caption} %
\usepackage{subcaption}
\usepackage[dvipsnames,table]{xcolor}
\usepackage{nicematrix}
\usepackage{tikz}
\usepackage{arydshln}
\usepackage{scalerel}
\usepackage[accsupp]{axessibility}  %
\usepackage{times} %
\usepackage[normalem]{ulem}

\usepackage{tabularray}
\UseTblrLibrary{booktabs}

\usetikzlibrary{positioning,fit,arrows.meta,shapes.geometric,intersections,calc,backgrounds}

\pgfplotsset{compat=1.18}

\NiceMatrixOptions
  {
    custom-line = 
     {
       letter = : ,
       command = dotrule , 
       ccommand = cdotrule ,
       tikz = {dotted, yshift=0.1pt}
     }
  }

\makeatletter
\renewcommand{\paragraph}{%
  \@startsection{paragraph}{4}%
  {\z@}{0.8ex \@plus 0ex \@minus .2ex}{-0.6em}%
  {\normalfont\normalsize\bfseries}%
}
\makeatother

\DeclareMathOperator*{\argmin}{arg\,min}

\newcommand{\xmark}{%
\tikz[scale=0.16,opacity=1] {
    \node[opacity=0, outer sep=0, inner sep=0, minimum width=0pt, minimum height=0pt] (text) at (0.5, 0.0) {\clap{\smash{\ding{55}}}};%
    \draw[line width=0.7,line cap=round] (0,0) to [bend left=6] (1,1);
    \draw[line width=0.7,line cap=round] (0.2,0.95) to [bend right=3] (0.8,0.05);
}}
\newcommand{\cmark}{%
\tikz[scale=0.17] {
    \node[opacity=0, outer sep=0, inner sep=0, minimum width=0pt, minimum height=0pt] (text) at (0.5, 0.0) {\clap{\smash{\ding{51}}}};%
    \draw[line width=0.7,line cap=round] (0.25,0) to [bend left=10] (1,1);%
    \draw[line width=0.7,line cap=round] (0,0.35) to [bend right=1] (0.23,0);%
}}

\newcommand{\PreserveBackslash}[1]{\let\temp=\\#1\let\\=\temp}
\newcolumntype{C}[1]{>{\PreserveBackslash\centering}p{#1}}
\newcolumntype{R}[1]{>{\PreserveBackslash\raggedleft}p{#1}}
\newcolumntype{L}[1]{>{\PreserveBackslash\raggedright}p{#1}}

\definecolor{Set2-0}{rgb}{0.398, 0.686, 0.314}
\definecolor{Set2-1}{rgb}{0.914, 0.616, 0.125}
\definecolor{Set2-2}{rgb}{0.125, 0.464, 0.788}
\definecolor{Set2-3}{rgb}{0.698, 0.133, 0.203}
\definecolor{Set2-4}{rgb}{0.635, 0.737, 0.271}
\definecolor{Set2-5}{rgb}{0.953, 0.224, 0.478}
\definecolor{Set2-6}{rgb}{0.260, 0.643, 0.796}
\definecolor{Set2-7}{rgb}{0.898, 0.734, 0.105}
\definecolor{gyellow}{RGB}{251,188,5}
\definecolor{gblue}{RGB}{66,133,244}
\definecolor{ggreen}{RGB}{52,168,83}
\definecolor{gred}{RGB}{234,67,53}
\definecolor{mymagenta}{RGB}{250,3,144}
\definecolor{myorange}{RGB}{232,157,65}

\definecolor{fireorange}{RGB}{255,115,2}  %
\definecolor{iceblue}{RGB}{175,227,255}  %

\colorlet{scene0color}{gyellow}
\colorlet{scene1color}{gblue}
\colorlet{scene2color}{ggreen}
\colorlet{scene3color}{gred}
\colorlet{scene4color}{mymagenta}
\colorlet{scene5color}{myorange}
\colorlet{singlescenecolor}{LimeGreen}

\colorlet{dinov2}{orange!30}
\colorlet{querybuffers}{Green!20}
\colorlet{mappingbuffers}{Salmon!30}
\colorlet{mapcodesgroup}{gray!20}
\colorlet{network}{Yellow!80!orange!30}
\colorlet{optimized}{fireorange}
\colorlet{frozen}{iceblue}

\colorlet{mlp}{gblue!50}
\colorlet{mha}{gyellow!50}

\colorlet{scenespecific}{gyellow!50}
\colorlet{sceneagnostic}{gblue!50}

\newcommand{\methodname}{\mbox{ACE-G}}

\def\radius{0.2}
\def\spacing{0.4} %
\def\rowspacing{1.4} %
\def\eps{0.01} %

\newcommand{\inlinebuffer}{
\protect\smash{\protect\scalerel*{
\protect\tikz[baseline]{
    \foreach \x in {0,1} {
        \foreach \y in {-1,0} {
            \pgfmathparse{60*rnd+40};
            \colorlet{mapcodecolor}{Gray!\pgfmathresult}
            \fill[mapcodecolor] (\x*\spacing-\eps, \y*\spacing+\eps) rectangle ++(\spacing+\eps,\spacing+\eps);
        }
    }
}}{\strut}}}

\newcommand{\inlinemapcode}{
\protect\smash{\protect\scalerel*{
\protect\tikz[baseline]{
    \foreach \x in {-1,0} {
        \foreach \y in {-1,0} {
            \pgfmathparse{60*rnd+40};
            \colorlet{mapcodecolor}{Gray!\pgfmathresult}
            \fill[mapcodecolor] (\x*\spacing, \y*\spacing) circle (\radius);
        }
    }
}}{\strut}}}

\pgfmathsetseed{2139012091}  %

\newcommand\blfootnote[1]{%
  \begingroup
  \renewcommand\thefootnote{}\footnote{#1}%
  \addtocounter{footnote}{-1}%
  \endgroup
}

\useunder{\uline}{\ul}{}

\definecolor{iccvblue}{rgb}{0.21,0.49,0.74}
\definecolor{gyellow}{RGB}{251,188,5}
\definecolor{gblue}{RGB}{66,133,244}
\definecolor{ggreen}{RGB}{52,168,83}

\usepackage[pagebackref,breaklinks,colorlinks,allcolors=iccvblue]{hyperref}

\def\paperID{8568} %
\def\confName{ICCV}
\def\confYear{2025}

\title{\methodname{}: Improving Generalization of Scene Coordinate Regression \\ Through Query Pre-Training}

\newcommand{\Hquad}{\hspace{0.6em}} %
\author{
Leonard Bruns\textsuperscript{2,*}\quad
Axel Barroso-Laguna\textsuperscript{1}\quad
Tommaso Cavallari\textsuperscript{1} \quad
{\'{A}}ron Monszpart\textsuperscript{4,*} \quad \\
Sowmya Munukutla\textsuperscript{1} \quad
Victor Adrian Prisacariu\textsuperscript{1,3} \quad
Eric Brachmann\textsuperscript{1}\\
\textsuperscript{1}Niantic Spatial \Hquad \textsuperscript{2}KTH Royal Institute of Technology \Hquad \textsuperscript{3}University of Oxford \Hquad \textsuperscript{4}Third Dimension AI\\\\
\url{https://nianticspatial.github.io/ace-g/}}

\begin{document}
\maketitle

\begin{abstract}
Scene coordinate regression (SCR) has established itself as a promising learning-based approach to visual relocalization. After mere minutes of scene-specific training, SCR models estimate camera poses of query images with high accuracy. Still, SCR methods fall short of the generalization capabilities of more classical feature-matching approaches. When imaging conditions of query images, such as lighting or viewpoint, are too different from the training views, SCR models fail. Failing to generalize is an inherent limitation of previous SCR frameworks, since their training objective is to encode the training views in the weights of the coordinate regressor itself. The regressor essentially overfits to the training views, by design. We propose to separate the coordinate regressor and the map representation into a generic transformer and a scene-specific map code. This separation allows us to pre-train the transformer on tens of thousands of scenes. More importantly, it allows us to train the transformer to generalize from mapping images to unseen query images during pre-training. We demonstrate on multiple challenging relocalization datasets that our method, \methodname{}, leads to significantly increased robustness while keeping the computational footprint attractive.
\end{abstract}

\blfootnote{\textsuperscript{*} Work done at Niantic.}

\begin{figure}
    \centering
    \includegraphics{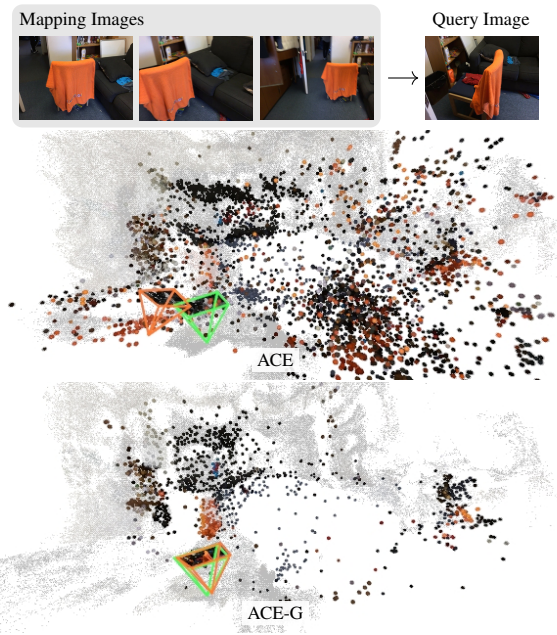}%
    \definecolor{figorange}{RGB}{254,126,50}%
    \definecolor{figgreen}{RGB}{74,254,90}%
    \caption{\methodname{} is trained with separate mapping and query splits explicitly optimizing scene coordinate regression for unseen views including changing conditions. The estimated scene coordinates for the query image and the \textcolor{figorange}{estimated} and \textcolor{figgreen}{ground-truth} camera poses are shown. \methodname{} estimates less noisy coordinates resulting in a more accurate pose estimate compared to ACE, which degrades for larger viewpoint or scene condition changes.}
    \label{fig:teaser}
    \vspace*{-\baselineskip}
\end{figure}

\section{Introduction}

The limits of our training data mean the limits of our world.
This statement, freely adapted from Wittgenstein, has been the guiding principle of machine learning and computer vision in recent years.
Scaling architectures and training data has led to astonishing successes in language models~\cite{achiam2023gpt}, image and video synthesis~\cite{videoworldsimulators2024}, and more recently 3D vision \cite{wang2024dust3r}.
Yet, there are particular tasks that are seemingly bound to small-scale learning problems.
One such task is scene coordinate regression (SCR) \cite{shotton2013scene}.

SCR has been proposed for the task of visual relocalization.
Given a set of RGB mapping images with known camera poses, one builds a visual map of an environment.
After the mapping stage, the system is presented with new query images from the same environment and asked to estimate their camera poses relative to the map.
SCR models solve this problem by training a scene-specific coordinate regressor on the mapping images.
The model learns to associate 2D pixels in the image with 3D coordinates in scene space.
Applied to a query image, the model's prediction induces 2D-3D correspondences between image and map that yield the desired camera pose.

Leading SCR frameworks, such as DSAC* \cite{brachmann2021visual} or ACE \cite{brachmann2023accelerated}, train scene-specific networks to encode the mapping images and to represent the map.
Their capability to generalize relies on the shift-invariance of their fully-convolutional networks, and on simplistic data augmentation, such as color jitter, applied to the mapping images \cite{brachmann2021visual}.
They excel in entirely static datasets, where there is no significant domain shift between mapping and query images \cite{brachmann2023accelerated}.
However, they struggle in more realistic scenarios when revisiting the same environment later, when lighting and other factors might have changed \cite{wang2024glace}.
We present a two-fold solution to the restricted generalization capabilities of previous SCR models. 

Firstly, we separate the map representation from the model predicting scene coordinates.
Our coordinate regressor is scene-agnostic. 
Its output depends on the query image, and a scene-specific map code.
The map code is our scene representation, kept separate from the coordinate regressor, and trained via back-propagation through the coordinate regressor at mapping time.
Keeping the coordinate regressor scene-agnostic enables pre-training it on large-scale data.
This allows us to switch to more powerful transformer architectures, such as ViT \cite{dosovitskiy2021an}, and to leverage more expressive features, such as DINO's \cite{oquab2024dinov}, without the danger of overfitting to scarce scene-specific data.

Secondly, we separately pre-train the coordinate regressor with mapping images as well as with query images.
When trained with mapping images, the regressor learns to produce sensible map codes that compress all scene-specific information.
When training with query images, we keep the map codes fixed, requiring the transformer to bridge any gap between mapping and query views.

Our method, \methodname{} (\cf\cref{fig:teaser}), effectively produces small map codes with fast mapping times, keeping the main advantages of previous SCR methods.
Extending on previous capabilities, our system learns to generalize to unseen image conditions via pre-training on 120\,000 mapping-query splits.
\methodname{} demonstrates superior robustness on two challenging indoor datasets with long term changes between mapping and query images.
We also show that the coordinate regressor generalizes well to environments outside of its training domain, making it practical and versatile.

\noindent \textbf{Our contributions:}
\begin{itemize}
    \item A framework, which we call \methodname{}, which combines a scene-agnostic coordinate regressor with scene-specific map codes. Map codes are few MB large, and trainable in minutes via back-propagation from posed RGB images.
    \item A pre-training procedure for \methodname{}. We intertwine training both map codes and the transformer from mapping views, then training the transformer with separate query views while keeping the map codes fixed. This encourages the transformer to learn to generalize.
    \item A scalable implementation that lets us pre-train on mapping-query splits of tens of thousands of scenes.%
\end{itemize}

\section{Related Work}

\paragraph{Visual Relocalization} has traditionally been solved via matching of sparse features \cite{Sattler12, sattler2016efficient, Lynen2019largescale, sarlin2019HFNet, compression2019cvpr, sarlin2021back}.
Posed mapping images are triangulated using structure-from-motion software \cite{schoenberger2016sfm,humenberger2020robust} to yield sparse point clouds where each 3D point is associated with high dimensional descriptors.
Features of the query image are matched to the point cloud resulting in 2D-3D correspondences.
Finally, the query camera pose is optimized using RANSAC \cite{fischler1981random} and PnP \cite{gao2003complete}.

Feature matching solutions still offer state-of-the-art accuracy and robustness when conditions between mapping and query images differ \cite{sattler2018aachendn2, jin2021image}.
The latest generation of feature matchers exhibit astonishing invariance to scale and viewpoint changes, as well as lighting changes up to day versus night \cite{detone2018superpoint, sarlin2020superglue, sun2021loftr, edstedt2023roma, edstedt2023dedode, barroso2024matching}.
These recent methods are learning-based and were trained on diverse image sets.

For example, MASt3R \cite{wang2024dust3r, leroy2024grounding}, a recent 3D foundation model showing state-of-the-art robustness, has been trained on a combination of multiple challenging datasets with hundreds of thousands of image pairs. 
While MASt3R is able to estimate metric-scale poses between pairs of images directly, its main operation mode for accurate visual relocalisation has been that of a feature matcher \cite{leroy2024grounding}.

The main disadvantage of feature-based relocalization is the considerable mapping time needed for triangulation, and the significant memory demand to store high dimensional descriptors for the scene point cloud.
While compression strategies have been proposed \cite{compression2019cvpr, Qunjie2022GoMatch}, they usually come with reduced performance on challenging datasets.

Absolute pose regression has been proposed as an alternative to feature-based relocalization \cite{kendall2015posenet}.
A neural network is trained on the mapping set to learn the association between  images and camera poses \cite{Shavit21multiscene, atloc, chen21}.
These methods initially suffered from low accuracy \cite{Sattler19} but recently improvements have been reported based on scene-specific data synthesis at the expense of very long mapping times \cite{Moreau21,chen2022dfnet}.
Relative pose regression networks learn to predict the relative pose between a query and a retrieved mapping image \cite{Balntas18, turkoglu2021visual, arnold2022map, dong2025reloc3r} or a panorama image capturing the scene \cite{zheng2025scene}. These networks can be scene-agnostic and pre-trained but still show comparatively low accuracy.

\paragraph{Scene Coordinate Regression (SCR)} is related to feature-based approaches but establishes 2D-3D correspondences via dense, direct regression.
Initially proposed for RGB-D inputs (using random forests \cite{shotton2013scene, valentin2015cvpr,cavallari2017fly}), SCR methods have later been adopted for RGB-only inputs, using neural networks \cite{brachmann2017dsac, Brachmann2018dsacpp, li2020hierarchical, brachmann2021visual, Wang2024hscnetpp, nguyen2024focustune, liu2024reprojection}.
In most works, the network is trained per scene to predict correspondences for that scene.
In terms of accuracy, SCR rivals feature-matching for small to medium sized scenes \cite{brachmann2021limits}, and some progress has been made towards larger environments \cite{Brachmann2019ESAC, wang2024glace}.

Initial incarnations of SCR required extensive training times on mapping data, similar or worse compared to the long mapping times of feature-based approaches.
ACE \cite{brachmann2023accelerated} introduced a training procedure to reduce mapping time to mere minutes, a speed-up making SCR even suitable for iterated training in a structure-from-motion setting \cite{brachmann2024acezero}.

Relatively few works aim at scene-agnostic coordinate regression. 
SANet \cite{Luwei2019sanet} and SACReg \cite{Revaud2024sacreg} have a scene-agnostic coordinate regressor that learns to interpolate 3D points of mapping images.
As such, they require an external 3D reconstruction or RGB-D data as a foundation. 
Marepo \cite{chen2024marepo} couples SCR with a scene-agnostic pose regressor.
Since the scene-specific SCR component is the bottleneck in terms of generalization, Marepo fails to outperform the SCR baseline in challenging situations.

Our work shares some conceptional similarity with NeuMap \cite{tang2023neumap}. Like us, they utilize a transformer coordinate regressor with scene-specific map codes. Their coordinate regressor is not fully scene-agnostic but trained per evaluation dataset and shared across those scenes. Training NeuMap further requires a full (sparse) 3D reconstruction for each evaluation scene. In contrast, our framework allows pre-training of a fully scene-agnostic coordinate regressor, and optimization of map codes from posed RGB images. Importantly, NeuMap misses any notion of query pre-training and aims primarily at scene compression at a dataset-level rather than generalization.

\section{Background}
Before diving into the architecture and training protocol we adopt for \methodname{}, we briefly summarize the main components of the Accelerated Coordinate Encoding (ACE) SCR.

The ACE scene coordinate regression model (\cf\cref{fig:hlarchitecture}, left) is formed by a scene-agnostic, fully-convolutional, image encoder and a scene-specific regression head (implemented as an MLP).
The image encoder maps image patches to high-dimensional feature vectors, and the regression head maps these features to scene coordinates.

ACE's training protocol involves first, as a preparation step, passing all mapping images through the image encoder in order to fill a training buffer with features for a large number of randomly sampled patches together with their corresponding metadata, that is, original 2D location in the input image, camera pose, and intrinsic parameters.

Then, in each training iteration, a random batch of $N$ features is sampled from 
the feature buffer and passed through the scene-specific MLP head to estimate the 3D scene points for the corresponding 2D pixels.

The ACE training loss used to update the weights of the head is then implemented as a pixel-wise projection loss applied to the 3D scene predictions in each training iteration. 
For further details, please refer to the original ACE paper~\cite{brachmann2023accelerated}.

\begin{figure}[tb!]
    \centering
    \begin{subfigure}[t]{0.35\linewidth}
        \centering
        \includegraphics{figures/flowchart_ace.tikz}
        \caption{ACE}
    \end{subfigure}%
    \begin{subfigure}[t]{0.5\linewidth}
        \centering
        \includegraphics{figures/flowchart_ours.tikz}
        \caption{\methodname{}}
    \end{subfigure}%
    \caption{\textbf{High-level comparison to ACE.} The \protect\tikz[baseline]{\protect\node[rounded corners=2pt, fill=scenespecific, inner sep=2pt,anchor=base] (A) {scene-specific};} MLP in ACE is replaced by a \protect\tikz[baseline]{\protect\node[rounded corners=2pt, fill=sceneagnostic, inner sep=2pt,anchor=base] (A) {scene-agnostic};} coordinate regressor and a scene-specific map code.}
    \label{fig:hlarchitecture}
\end{figure}
\begin{figure*}[tb!]
    \centering
    \begin{subfigure}{0.49\linewidth}
        \centering%
        \includegraphics{figures/overview_mapping_training.tikz}%
        \caption{Pre-Training (\cref{sec:multiscene}): Mapping Iteration}
    \end{subfigure}
    \begin{subfigure}{0.49\linewidth}
        \centering%
        \includegraphics{figures/overview_query_training.tikz}%
        \caption{Pre-Training (\cref{sec:multiscene}): Query Iteration}
    \end{subfigure}
    \begin{subfigure}{0.49\linewidth}
        \centering%
        \includegraphics{figures/overview_novel_scene.tikz}%
        \caption{Novel Scene Mapping (\cref{sec:singlescene})}
    \end{subfigure}
    \begin{subfigure}{0.49\linewidth}
        \centering%
        \includegraphics{figures/overview_novel_image.tikz}%
        \caption{Image Relocalization (\cref{sec:relocalization})}
    \end{subfigure}
    \caption{\textbf{Overview of \methodname{}}. The \protect\tikz[baseline]{\protect\node[rounded corners=2pt, fill=network, inner sep=2pt,anchor=base] (A) {scene-agnostic coordinate regressor $f_\theta$};} is pre-trained by alternating between mapping iterations (a) and query iterations (b).\enspace (a) During pre-training mapping iterations, both map codes and network weights are \protect\tikz[baseline]{\protect\node[rounded corners=2pt, draw=fireorange, text depth=1pt, thick, inner sep=2pt,anchor=base] (A) {optimized};} \protect\smash{\protect\scalerel*{\includegraphics{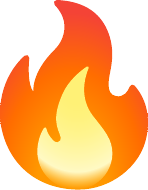}}{\strut}} using precomputed buffers \protect\inlinebuffer\, storing shuffled \protect\tikz[baseline]{\protect\node[rounded corners=2pt, fill=dinov2, inner sep=2pt,anchor=base] (A) {DINOv2};} features and meta data necessary for supervision (\cf \cite{brachmann2023accelerated}). For each mapping buffer a corresponding map code \protect\inlinemapcode\, is optimized. \enspace (b) During pre-training query iterations, the map codes are \protect\tikz[baseline]{\protect\node[rounded corners=2pt, draw=frozen, text depth=1pt, thick, inner sep=2pt,anchor=base] (A) {frozen};} \protect\smash{\protect\scalerel*{\includegraphics{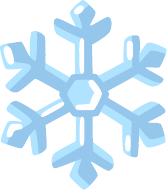}}{\strut}} and the network is trained to estimate scene coordinates for query buffers made up of viewpoints or scene conditions different from the mapping buffers'. \enspace (c) Once the regressor has been pre-trained, a novel scene can be encoded in a new map code by minimizing the reprojection error. \enspace (d) Given such an optimized map code and a new query image, scene coordinates and uncertainty can be estimated via a forward pass. The resulting 2D-3D correspondences can then be used to estimate the camera pose.}
    \label{fig:overview}
    \vspace{-1em}
\end{figure*}

\section{Method}

We propose to replace the scene-specific MLP used in ACE with a scene-agnostic coordinate regressor that takes as input a latent map code and an image patch embedding and estimates the corresponding scene coordinate.
See~\cref{fig:hlarchitecture} for a high-level description of the architecture.
A map code describes the feature $\to$ coordinate mapping for a scene. Specifically, let a map code $\mathcal{C}=\{ \boldsymbol{c}_i \in \mathbb{R}^{D_\mathrm{map}} \mid i = 1, \dots, N_\mathcal{C} \}$ denote a scene-specific set of $D_\mathrm{map}$-dimensional map embeddings, $\boldsymbol{e}\in \mathbb{R}^{D_\mathrm{feat}}$ a $D_\mathrm{feat}$-dimensional patch embedding; and $\boldsymbol{y}\in\mathbb{R}^3$ the estimated scene coordinate. Our network can then be formally defined as
\begin{equation}\label{eq:function}
    \begin{aligned}
        f_\theta: \mathbb{R}^{D_\mathrm{feat}} \times \mathcal{P}\left(\mathbb{R}^{D_\mathrm{map}}\right) &\to \mathbb{R}^3 \\
        (\boldsymbol{e}, \mathcal{C}) &\mapsto \boldsymbol{y},
    \end{aligned}
\end{equation}
where $\mathcal{P}(\cdot)$ denotes the power set, and $\theta$ the parameters.

\Cref{fig:overview} provides an overview of the different stages of optimization and inference.
During pre-training of the scene-agnostic regressor (\cref{sec:multiscene}), the network is simultaneously optimized across multiple scenes using disjoint sets of mapping and query images.
The mapping images are used to infer scene-specific map codes, while the query images are used to optimize the scene-agnostic network aiming to improve generalization for novel views, given the previously optimized map codes.
Once the scene-agnostic coordinate regressor has been pre-trained, the latent map code for a new scene can be optimized within minutes (\cref{sec:singlescene}), akin to the per-scene optimization in ACE.
To relocalize a query image, the scene-specific map code is used to condition the prediction of the 3D coordinates for all image patches, and PnP with RANSAC can then be used to estimate the camera pose (\cref{sec:relocalization}).

\subsection{Architecture}\label{sec:architecture}

At a high-level, the architecture of \methodname{} consists of a pre-trained, generic, image encoder and a scene-agnostic coordinate regressor (\cref{fig:hlarchitecture}).

\paragraph{Image Encoder} As the image encoder we use DINOv2 \cite{oquab2024dinov,darcet2024vision} hypothesizing that higher-level image understanding embedded in the resulting features might be beneficial for more difficult query images (\cf \eg, \cite{barroso2024matching}).
Given an input image $\mathbf{I}$, the encoder $f_\mathrm{enc}$ predicts a set of patch embeddings $\mathcal{E}=\{\boldsymbol{e}_i \in \mathbb{R}^{D_\mathrm{feat}} \mid i=1,\dots,N_\mathcal{E}\}$.

\paragraph{Scene-Agnostic Coordinate Regressor}

\begin{figure}[htb]
    \centering
    \includegraphics{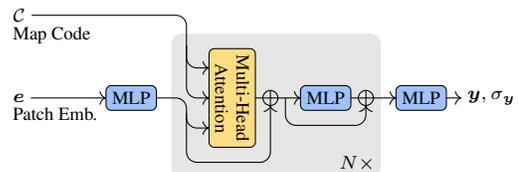}
    \caption{\textbf{Network architecture.} The scene-agnostic coordinate regressor consists of $N$ cross-attention-only blocks. Given a patch embedding $\boldsymbol{e}$ and a map code $\mathcal{C}$ it estimates the 3D scene coordinate $\boldsymbol{y}$ and uncertainty $\sigma_{\boldsymbol{y}}$.}\label{fig:network}
    \vspace*{-\baselineskip}%
\end{figure}

Our scene-agnostic coordinate regressor is shown in \cref{fig:network}. It consists of $N$ cross-attention-only vision transformer blocks \cite{dosovitskiy2021an}. Patch embeddings $\boldsymbol{e}$ are used as query tokens and the individual map embeddings in the map code $\mathcal{C}$ are used as key and value tokens. The map embeddings use no additional positional encodings and are therefore seen as permutation-invariant by the transformer. Following the transformer, a 2-layer MLP without layer normalization \cite{ba2016layer} is used to regress the scene coordinate $\boldsymbol{y}$ and standard deviation $\sigma_{\boldsymbol{y}}$ of a Laplace distribution (see loss below).

\subsection{Mapping/Query Pre-Training}\label{sec:multiscene}

Our goal is to train a scene-agnostic coordinate regressor which, after finding the scene-specific map code, generalizes further than simply optimizing an MLP for the same training data. To achieve this, we train the scene-agnostic coordinate regressor with query images separate from the mapping images that were used to find the map codes. This is akin to how the network is later used to map novel scenes and relocalize query images: the map code is inferred from one sequence of images (\ie, the mapping images), and a query image taken potentially during a different day or from a different viewing angle is to be localized relative to the mapping images given only the map code.

For pre-training we consider datasets for which ground-truth scene coordinates are available. Either from an RGB-D sensor, synthetic data generation, or generated from RGB through multi-view stereo (\cf \cref{tab:datasets}). Note that our architecture does not require ground-truth scene coordinates when optimizing map codes on novel scenes (see \cref{sec:singlescene}). 

\paragraph{Multi-Buffer Dataset} Previously, it has been shown that precomputing image features and storing them as a shuffled buffer can significantly speed up the optimization of scene coordinate regressors as repeated feature extraction is avoided and patches from all views are shuffled, de-correlating the gradients within each batch \cite{brachmann2023accelerated}.
Following this approach, we precompute and store the buffers for all mapping and query training sequences. %

Let $\mathcal{B}=\{ (\mathcal{M}_i,\mathcal{Q}_i) \mid i = 1,\dots,N_\mathcal{B}) \}$ denote the dataset of $N_\mathcal{B}$ mapping-query tuples. Each tuple $i$ consists of a mapping buffer $\mathcal{M}_i=\{ (\boldsymbol{e}_j, \boldsymbol{y}_j) \mid j= 1,\dots,M_i \}$ and query buffer $\mathcal{Q}_i=\{ (\boldsymbol{e}_j, \boldsymbol{y}_j) \mid j=1,\dots,Q_i \}$ made up of patch embeddings $\boldsymbol{e}_j$ and corresponding ground-truth scene coordinates $\boldsymbol{y}_j$. During the optimization, each mapping-query tuple $i$ is associated with a map code $\mathcal{C}_i\in \mathcal{G}$\footnote{Here, $\mathcal{G}=\{\mathcal{C}_i\mid i=1,\dots,N_\mathcal{B}\}$ denotes a set of map codes $\mathcal{C}_i.$} that will be optimized on the corresponding mapping buffer $\mathcal{M}_i$.

\paragraph{Mapping and Query Optimization}
Our pre-training alternates between ``mapping'' and ``query'' optimization iterations.
During mapping iterations, the scene-agnostic network parameters and map codes are optimized jointly:
\begin{equation}\label{eq:mapping}
    \theta^\ast, \mathcal{G}^\ast = \argmin_{\theta,\mathcal{G}}\mathbb{E}_{\mathcal{M}}\left[ 
        \log \tilde{\sigma}_{\boldsymbol{y}} + \sqrt{2}\frac{\lVert \tilde{\boldsymbol{y}} - \boldsymbol{y} \rVert }{\tilde{\sigma}_{\boldsymbol{y}}}
    \right].
\end{equation}
Here, $(\tilde{y},\tilde{\sigma}_{\boldsymbol{y}})=f_\theta(\boldsymbol{e}, \mathcal{C}_i)$ denote the estimates for a patch embedding $\boldsymbol{e}$ with ground-truth coordinate $\boldsymbol{y}$ sampled from a mapping buffer $\mathcal{M}_i\sim\mathcal{M}$.
Intuitively, this stage ensures that the network has the capacity to fit various types of scenes and move information from the buffers into the map codes.

Conversely, during query iterations, only the scene-agnostic network parameters are updated, keeping the map codes $\mathcal{C}^\ast$ fixed:
\begin{equation}\label{eq:query}
    \theta^\ast = \argmin_\theta \mathbb{E}_{\mathcal{Q}}\left[  
        \log \tilde{\sigma}_{\boldsymbol{y}} + \sqrt{2}\frac{\lVert \tilde{\boldsymbol{y}} - \boldsymbol{y} \rVert }{\tilde{\sigma}_{\boldsymbol{y}}} 
    \right],
\end{equation}
where $(\tilde{y},\tilde{\sigma}_{\boldsymbol{y}})=f_\theta(\boldsymbol{e}, \mathcal{C}^\ast_i)$.
This stage aims to improve generalization beyond previous scene coordinate regressors that are optimized just on mapping images.

For both mapping and query iterations, our optimization objective is the negative log-likelihood of the ground-truth scene coordinates under an estimated Laplace distribution. 
That is, after pre-training, $\tilde{\sigma}_{\boldsymbol{y}}$ can be interpreted as the estimated standard deviation of the scene coordinate in 3D.

\paragraph{Implementation} Instead of optimizing the network and map codes using all scenes simultaneously until convergence, similar to an autodecoder \cite{bojanowski2018optimizing, park2019deepsdf,tang2023neumap}, we opt to repeatedly optimize scenes from scratch throughout the training process.
Specifically, we only maintain a subset of $N_\mathrm{active}$ scenes $\mathcal{B}_\mathrm{active} \subset \mathcal{B}$ and a map code for each.
Each of the active map codes is updated until a randomized number of iterations is reached, at which point it is reset (we initialize map codes by sampling from a Gaussian distribution with $\sigma=0.01$) and a new scene configuration is sampled from $\mathcal{B}$ and added to the pool, replacing the previous one.

During pre-training, we iteratively perform mapping iterations followed by query iterations.
Mapping iterations optimize both the network parameters and the map codes, using \cref{eq:mapping} and batches sampled from the $\mathcal{M}$ buffers.
Query iterations then use \cref{eq:query}, keeping the map codes frozen and sampling batches from the $\mathcal{Q}$ buffers. To reduce the chance of overfitting the network to the currently active scenes, we further only perform network updates in every 10th iteration, that is, only the map codes are updated in 9 out of 10 optimization iterations. Further, during query optimization we skip scenes when their corresponding map code has undergone less than $N_\mathrm{qstandby}$ mapping iterations.
Intuitively, we want the scene-agnostic network to learn generalizing from map codes that are well initialized, instead of spending capacity on estimating the uncertainty for insufficiently optimized map codes.
Each iteration is based on randomly sampled batches of the active scenes' buffers composed of $N_\mathrm{spb}$ scenes per batch and $N_\mathrm{pps}$ patches per scene.

\subsection{Novel Scene Mapping}\label{sec:singlescene}

To find the map code for a new scene we mainly follow the approach used by ACE \cite{brachmann2023accelerated}:
given a new set of posed images $\{\mathbf{I}_i, \tensor*[^{\mathrm{w}}]{\mathbf{T}}{_{\mathrm{c}i}} \mid i=1,\dots,N\}$, the mapping buffer $\mathcal{M}$ is prepared ahead of training.
However, instead of the ground-truth scene coordinate $\boldsymbol{y}$, the image coordinate $\boldsymbol{x}$, camera pose $\tensor*[^{\mathrm{w}}]{\mathbf{T}}{_{\mathrm{c}}}$, and intrinsic matrix $\mathbf{K}$ are stored for each patch embedding $\boldsymbol{e}$.
The map code for a novel scene is then optimized by minimizing the negative log-likelihood in 2D:
\begin{equation}
    \mathcal{C}^\ast = \argmin_{\mathcal{C}}\mathbb{E}_{\mathcal{M}}\left[ 
        \log \tilde{\sigma}_{\boldsymbol{x}} + \sqrt{2}\frac{\lVert \tilde{\boldsymbol{x}} - \boldsymbol{x} \rVert }{\tilde{\sigma}_{\boldsymbol{x}}}
    \right],
\end{equation}
where $\tilde{\boldsymbol{x}}$ and $\tilde{\sigma}_{\boldsymbol{x}}$ are derived from $(\tilde{\boldsymbol{y}},\tilde{\sigma}_{\boldsymbol{y}})=f_\theta(\boldsymbol{e}, \mathcal{C})$ via pinhole projection. For invalid estimates (behind the camera or with large reprojection errors), the constant depth prior of ACE is used but modified to take uncertainty into account, similarly to \cref{eq:mapping}.

\begin{table*}[tbh]
    \centering
    \caption{\textbf{Results on Indoor-6.} Each cell contains in order: percent of correctly localized frames under $(5\si{\degree}, 25\,\mathrm{cm})$, median translation error in $\mathrm{cm}$, and median rotation error in degrees. Best in each SCR group \textbf{highlighted}.}
    \vspace{-0.5\baselineskip}
    \renewcommand{\arraystretch}{1.1}
    \setlength{\tabcolsep}{2pt}
    \footnotesize
    {\begin{NiceTabular}{@{}llccccccccc@{}}
    \toprule
        & \multirow{2}{*}{\quad$(\%\,/\,\mathrm{cm}\,/\,\si{\degree})$}& \multirow{2}{*}{\parbox{1cm}{\centering Map.\\Time}} & \multirow{2}{*}{\parbox{1cm}{\centering Map Size}} & \multicolumn{7}{c}{Indoor-6}\\
        \cmidrule(lr){5-11}
        & & & & Scene 1\ & Scene 2a & Scene 3 & Scene 4a & Scene 5 & Scene 6 & Avg.  \\
    \midrule
        \multirow{2}{*}{\rotatebox{90}{\parbox{7mm}{\centering\scriptsize Non\\SCR}}} &  Reloc3r \cite{dong2025reloc3r} & \cellcolor{green!30} $<$ 5 min. & \cellcolor{red!70!Dandelion!30}$\sim$500 MB & 76 / 8 / 0.6 & 76 / 10 / 0.9 & 87 / 4 / 0.5 & 82 / 9 / 0.8 & 50 / 22 / 1.4 & 86 / 3 / 0.6 & 76 / 9 / 0.8 \\ 
        & MASt3R+Kapture \cite{leroy2024grounding} & \cellcolor{red!30}$ $5-10 h & \cellcolor{red!30}$\sim$5 GB & 92 / 2.4 / 0.4 & 97 / 2.6 / 0.3 & 96 / 1.8 / 0.4 & 96 / 2.6 / 0.5 & 85 / 4.3 / 0.7 & 95 / 1.3 / 0.3 & 94 / 2.5 / 0.5 \\
    \midrule
        \multirow{2}{*}{\rotatebox{90}{\parbox{7.2mm}{\centering\scriptsize SCR\\\rlap{25 min.}}}}  & GLACE \cite{wang2024glace} & \cellcolor{Dandelion!50}25 min. & \cellcolor{green!30} 6 MB & 90 / 3.8 / 0.7 & \textbf{100} / \textbf{3.9} / \textbf{0.4} & 93 / \textbf{2.9} / \textbf{0.6} & \textbf{99} / \textbf{2.3} / \textbf{0.5} & 79 / 6.1 / 0.9 & \textbf{97} / \textbf{2.1} / \textbf{0.4} & 93 / \textbf{3.5} / \textbf{0.6} \\
        & Ours (25 min.) & \cellcolor{Dandelion!50}25 min. & \cellcolor{Green!90!Dandelion!30} 12 MB & \textbf{96} / \textbf{3.5} / \textbf{0.6} & 98 / \textbf{3.9} / \textbf{0.4} & \textbf{98} / 3.8 / 0.8 & \textbf{99} / 3.2 / 0.8 & \textbf{97} / \textbf{5.1} / \textbf{0.8} & 91 / 2.7 / 0.6 & \textbf{96} / 3.7 / 0.7 \\[\aboverulesep]
\dotrule \noalign{\vskip \belowrulesep}
        \multirow{3}{*}{\rotatebox{90}{\parbox{1cm}{\centering\scriptsize SCR\\5 min.}}} & ACE \cite{brachmann2023accelerated} & \cellcolor{green!30}5 min. & \cellcolor{green!30} 4 MB & 52 / 17.7 / 2.4 & 86 / 8.0 / 0.8 & 66 / 9.0 / 1.6 & 94 / 4.6 / 0.9 & 51 / 21.6 / 4.0 & 70 / 5.2 / 0.9 & 70 / 11.0 / 1.8 \\
        & DINO-ACE & \cellcolor{green!30}5 min. & \cellcolor{green!30} 4 MB & 90 / 5.4 / 0.9 & 97 / 5.2 / 0.6 & 91 / 6.0 / 1.2 & 96 / 4.7 / 1.2 & 89 / 7.5 / 1.2 & 85 / 4.8 / 1.0 & 91 / 5.6 / 1.0 \\
        & Ours (5 min.) & \cellcolor{green!30}5 min. & \cellcolor{Green!90!Dandelion!30} 12 MB & \textbf{94} / \textbf{4.3} / \textbf{0.7} & \textbf{98} / \textbf{4.8} / \textbf{0.4} & \textbf{96} / \textbf{4.7} / \textbf{0.9} & \textbf{98} / \textbf{3.7} / \textbf{0.9} & \textbf{93} / \textbf{5.9} / \textbf{1.0} & \textbf{92} / \textbf{3.9} / \textbf{0.8} & \textbf{95} / \textbf{4.5} / \textbf{0.8} \\
    \bottomrule
    \end{NiceTabular}}
    \label{tab:indoor6}
    \vspace*{-\baselineskip}
\end{table*}

\subsection{Image Relocalization}\label{sec:relocalization}

Given a new image $\mathbf{I}$ for a previously mapped scene with map code $\mathcal{C}^\ast$, the scene coordinates are estimated using a simple forward pass through the image encoder and the scene-agnostic coordinate regressor.
Specifically, each image patch (with its known 2D position $\boldsymbol{x}_i$) is first mapped to a patch embedding $\boldsymbol{e}_i\in \mathcal{E} = f_\mathrm{enc}(\mathbf{I})$ via the image encoder and passed to the pre-trained scene-agnostic regressor to estimate the corresponding 3D coordinate and uncertainty $\tilde{\boldsymbol{y}}_i,\tilde{\sigma}_{\boldsymbol{y},i}=f_\theta(\boldsymbol{e}_i, \mathcal{C}^\ast)$.
The resulting set of 2D-3D correspondences $\{(\boldsymbol{x}_i,\tilde{\boldsymbol{y}}_i) \mid i = 1,\dots,N_\mathrm{patches}\}$ can directly be used with PnP and RANSAC to estimate the camera pose $\tensor*[^{\mathrm{w}}]{\widetilde{\mathbf{T}}}{_{\mathrm{c}}}$ (see \cite{brachmann2021visual} for further details on camera pose estimation from the set of 2D-3D correspondences).

We find that the estimated uncertainty $\tilde{\sigma}_{\boldsymbol{y}}$ can be used to prefilter the correspondences fed to the RANSAC algorithm.
Specifically, we use an adaptive uncertainty threshold based on the lowest $p$-quantile of uncertainties, that is, $\sigma_\mathrm{thresh}= f\cdot  Q_p(\Sigma) $, where $Q_p$ denotes the $p$-quantile and $\Sigma$ is the set of estimated uncertainties. A reduced set of correspondences $\{(\boldsymbol{x}_i,\tilde{\boldsymbol{y}}_i) \mid i = 1,\dots,N_\mathrm{patches} \land \tilde{\sigma}_{\boldsymbol{y},i}<\sigma_\mathrm{thresh}\}$ is then used to estimate the camera pose. In all experiments we use $p=0.1$ and $f=2.0$.

\begin{table}[tb]
    \centering
    \caption{\textbf{Training datasets.} (\# scenes: number of scans included in the training; Cov.: uses image pairs by \cite{wang2024dust3r}; Sep.\ Q.: separate query sequence; 3D: source of 3D information)}%
    \vspace{-0.5\baselineskip}%
    \renewcommand{\arraystretch}{1.1}%
    \setlength{\tabcolsep}{2pt}%
    \footnotesize%
    \begin{tabular}{@{} l r c c c c @{}}
        \toprule
        & \# Scenes & Cov. & Sep. Q. & Metric & 3D\\
        \midrule
        ScanNet \cite{dai2017scannet} & 1201 & \text{\protect\xmark} & \text{\protect{\xmark}} & \text{\protect{\cmark}}  & RGB-D \\
        ScanNet++ v1 \cite{yeshwanth2023scannet++} & 199 & \text{\protect\cmark} & \text{\protect{\xmark}} & \text{\protect{\cmark}}  & RGB-D \\
        ARKitScenes \cite{baruch2021arkitscenes} & 4520 & \text{\protect\cmark} & \text{\protect{\xmark}} & \text{\protect{\cmark}}  & RGB-D \\
        MapFree \cite{arnold2022map} & $2 \times 400$ & \text{\protect\xmark} & \text{\protect{\cmark}} & \text{\protect{\cmark}} & MVS \cite{bleyer2011patchmatch} \\
        BlendedMVS \cite{yao2020blendedmvs} & 462 & \text{\protect\xmark} & \text{\protect{\xmark}} & \text{\protect{\xmark}} & Synth. \\
        WildRGBD \cite{xia2024rgbd} & 22359 & \text{\protect\xmark} & \text{\protect{\xmark}} & \text{\protect{\cmark}} & RGB-D \\
        \bottomrule
    \end{tabular}
    \label{tab:datasets}
    \vspace*{-1\baselineskip}
\end{table}

\section{Experiments}

\paragraph{Pre-Training Datasets} \Cref{tab:datasets} provides an overview of the datasets used to pre-train the scene-agnostic coordinate regressor.
For all datasets, for each scene, we select ``mapping'' and ``query'' chunks by splitting  the sequences into disjoint -- likely solvable -- mapping and query sections. 
The distributions are adjusted per dataset to account for differences in camera trajectories and scene content.
For some datasets, the information about co-visible training image pairs published by \cite{wang2024dust3r} is used to inform the sampling procedure.
For each dataset we sample 20\,000 map/query configurations, then pre-compute a resulting total of 240\,000 $\mathcal{M}$ and $\mathcal{Q}$ buffers. See supplementary material for further information and qualitative training data samples.

\paragraph{Evaluation Datasets} We focus our evaluation on %
challenging datasets 
with varying lighting conditions, strong view point differences, and long-term changes.
We report the performance of \methodname{} on the Indoor-6 \cite{do2022scenelandmarkloc}, RIO10 \cite{wald2020beyond}, and Cambridge Landmarks \cite{kendall2015posenet} datasets.
Indoor-6 depicts large multi-room indoor environments, including significant lighting variations between mapping and query scans.
RIO10 challenges the algorithms with scans captured over the course of several weeks and months, showing significant day-to-day variations in the positions of objects between scans, in addition to variable lighting conditions.
For this dataset we train the map codes on the ``mapping'' scans and compute the error metrics on the ``validation'' scans.
Cambridge Landmarks is used to assess the performance in larger outdoor environments, and puts to the test the capabilities of \methodname{} as none of the training datasets contain similarly large, outdoor environments. %

\begin{table*}[htb!]
    \centering
    \caption{\textbf{Results on RIO10.} Each cell contains: median translation error and median rotation error. Best in each SCR group \textbf{highlighted}.}
    \vspace{-0.5\baselineskip}
    \renewcommand{\arraystretch}{1.1}
    \setlength{\tabcolsep}{2pt}
    \scriptsize
    \begin{NiceTabular}{@{}lccccccccccc@{}}
    \toprule
         \quad$(\mathrm{cm}\,/\,\si{\degree})$ & Scene 1 & Scene 2 & Scene 3 & Scene 4 & Scene 5 & Scene 6 & Scene 7 & Scene 8 & Scene 9 & Scene 10 & Avg. \\
    \midrule
        Reloc3r \cite{dong2025reloc3r} & 29 / 3.8 & 23 / 3.2 & 45 / 4.1 & 86 / 53.4 & 35 / 3.8 & 10 / 2.1 & 20 / 2.5 & 95 / 7.7 & 76 / 7.4 & 51 / 6.3 & 47 / 9.4 \\
        MASt3R+Kapture \cite{leroy2024grounding} &  69 / 26.4 & 44 / 14.9 & 61 / 22.2 &N/A & 68 / 22.9 & 7.0 / 2.7 & 8.9 / 2.9 & N/A & N/A & 106 / 46.9 & N/A \\
    \midrule
        GLACE \cite{wang2024glace} & 22.1 / 7.8 & \textbf{21} / \textbf{6.1} & \textbf{34} / \textbf{10.5} & 266.2 / 77.9 & 127.8 / 45.6 & 18.8 / 6.9 & \textbf{19.9} / \textbf{5.2} & 221.1 / 53.4 & 323.3 / 89.8 & 85.5 / 30.9 & 114 / 33.4 \\
        Ours (25 min.) & \textbf{20.9} / \textbf{6.8} & 25.8 / 7.6 & 40.2 / 10.9 & \textbf{88.2} / \textbf{32.9} & \textbf{17.6} / \textbf{6.4} & \textbf{14.4} / \textbf{4.7} & 26.5 / 8.6 & \textbf{68.3} / \textbf{16.7} & \textbf{29.2} / \textbf{9.6} & \textbf{39.1} / \textbf{17.3} & \textbf{37.0} / \textbf{12.1} \\[\aboverulesep]
\dotrule \noalign{\vskip \belowrulesep}
        ACE \cite{brachmann2023accelerated} &  97.1 / 25.8 & 95.1 / 22.6 & 159.8 / 35.1 & 390.3 / 91.2 & 915.9 / 95.0 & 170.5 / 57.1 & 57.7 / 13.5 & 481.8 / 57.1 & 361.6 / 100.3 & 854.8 / 89.2 & 358.4 / 58.7 \\
        DINO-ACE &  48.3 / 13.0 & 45.1 / 11.6 & 63.5 / 13.2 & 108.9 / \textbf{32.7} & 118.2 / 10.6 & 21.6 / 7.0 & 53.0 / 12.8 & 233.9 / 21.2 & 53.6 / 16.9 & 92.1 / 20.2 & 83.8 / 15.9 \\
        Ours (5 min.)& \textbf{25.2} / \textbf{8.3} & \textbf{29.7} / \textbf{8.5} & \textbf{45.6} / \textbf{12.4} & \textbf{89.3} / 37.8 & \textbf{20.3} / \textbf{7.4} & \textbf{17.8} / \textbf{5.6} & \textbf{30} / \textbf{9.7} & \textbf{75.3} / \textbf{18.7} & \textbf{32.9} / \textbf{11.0} & \textbf{44.5} / \textbf{18.0} & \textbf{41.1} / \textbf{13.8} \\
    \bottomrule
    \end{NiceTabular}%
    \label{tab:rio10}%
    \vspace*{-\baselineskip}%
\end{table*}

\paragraph{Baselines} We compare \methodname{} with five primary baselines: ACE \cite{brachmann2023accelerated} in its default setting, using its fully-convolutional encoder; a modified version of ACE, ``DINO-ACE'', where we replaced the encoder with DINOv2 \cite{oquab2024dinov}, \ie, the same encoder that we are using; GLACE \cite{wang2024glace}, which improves over ACE by adding a noise-augmented global encoding among other modifications; Reloc3r \cite{dong2025reloc3r}, and MASt3R \cite{leroy2024grounding} which both rely on image retrieval and -- just for the latter -- SfM models generated via Kapture \cite{humenberger2020robust} for map-relative localization \cite{wang2024dust3r}. Reliance on SfM reconstructions leads to significant mapping times and memory demands for MASt3R. Reloc3r maps quickly by building a retrieval index, but needs to store the mapping images.
We present results for two variants of \methodname{}: the default one that can perform mapping of novel scenes in $\approx$5 min on a single GPU, similarly to the default configuration of ACE; and one configuration tuned to optimize map codes in $\approx$25 min, to be fairly comparable with GLACE. ACE-G maps consume 12 MB of memory stored using full precision ($N_\mathcal{C}=4096, D_\mathrm{map}=768$).

\paragraph{Metrics} We report the median translation and rotation error of the estimated poses, as well as the percentage of correctly localized frames under different translation and rotation error thresholds.

\paragraph{Hyperparameters} We pre-train for 4.4M iterations on 8 A100 GPUs with a scene batch size $N_\mathrm{spb}=200$ and patch batch size $N_\mathrm{pps}=512$. Each map code is optimized between 6\,000 and 10\,000 iterations with $N_\mathrm{qstandby}=5000$. See supplementary material for further details.

\begin{figure}[tb]
    \centering
    \begin{subfigure}{0.49\linewidth}
        \includegraphics[width=\linewidth,trim={5cm 1cm 7cm 2cm},clip]{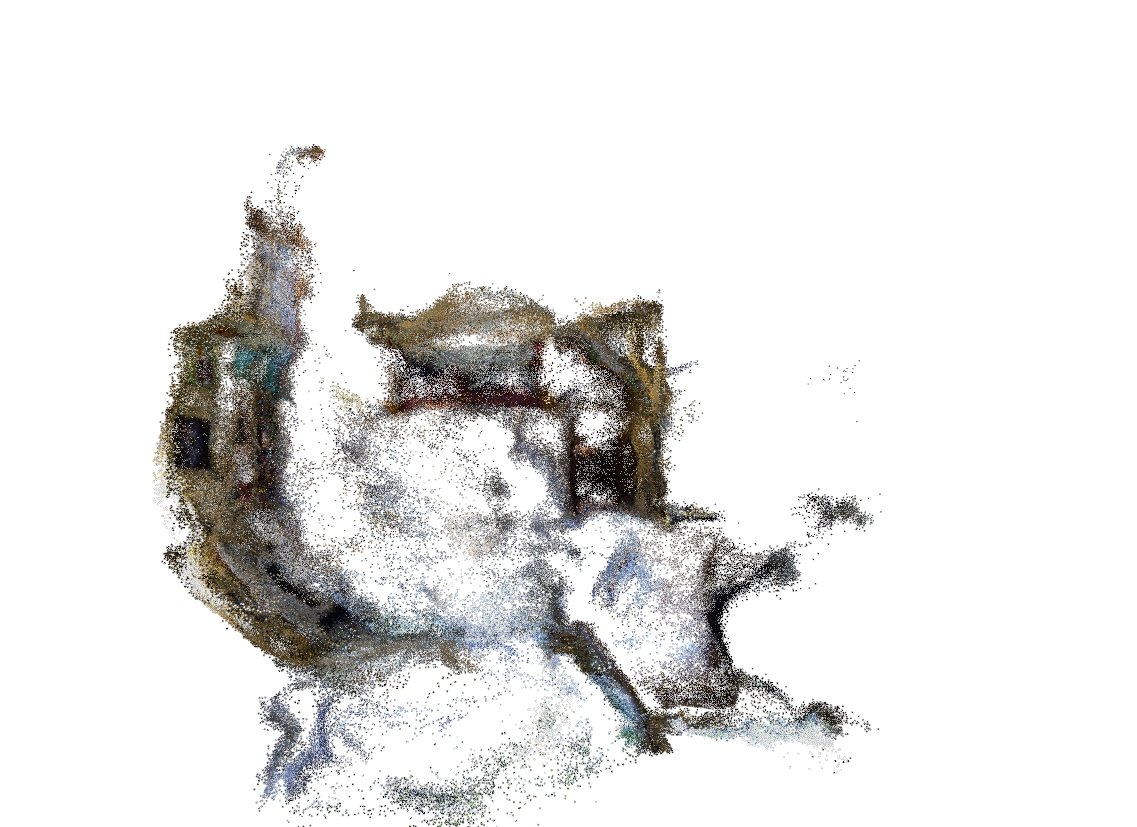}
    \end{subfigure}
    \begin{subfigure}{0.49\linewidth}
        \includegraphics[width=\linewidth,trim={5cm 1cm 7cm 2cm},clip]{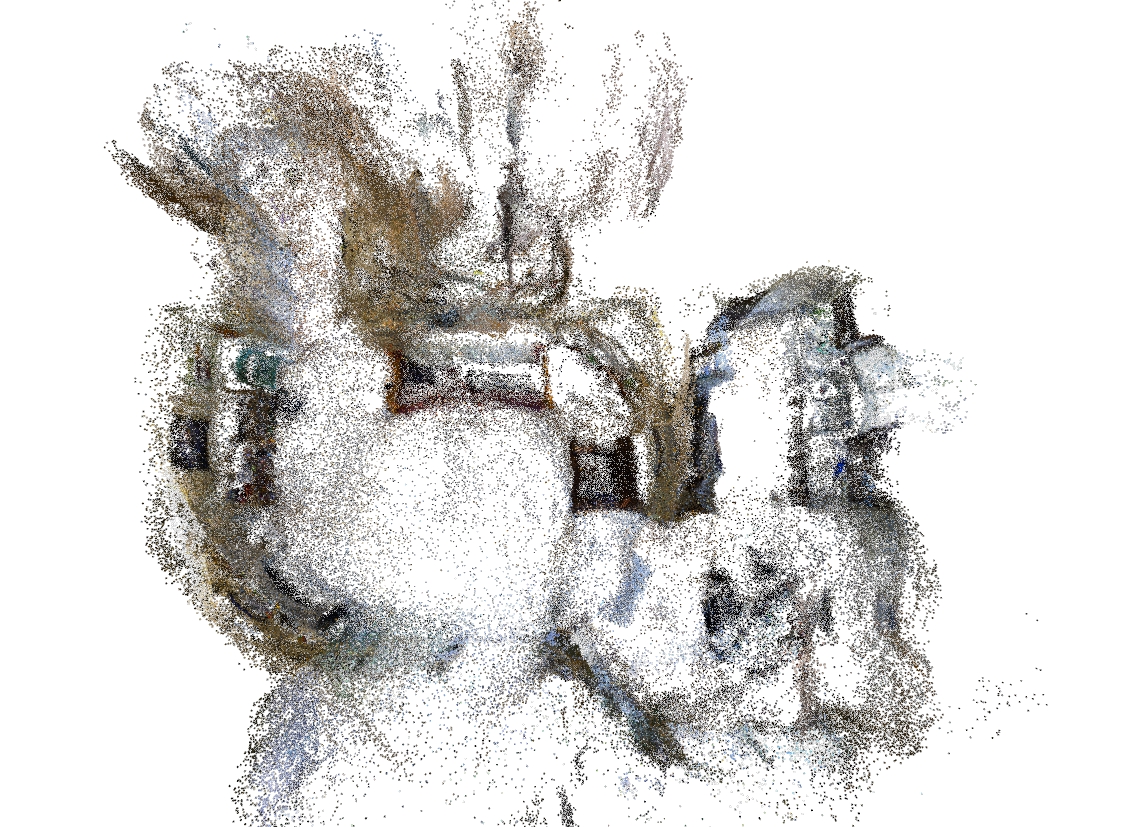}
    \end{subfigure}
    \caption{\textbf{Scene reconstruction.} Comparing the estimated scene coordinates of ACE (left) and \methodname{} (right) on Scene 1 of Indoor-6 shows that ACE fails to reconstruct some parts of the apartment.}%
    \label{fig:scenecoords}%
    \vspace*{-\baselineskip}%
\end{figure}

\subsection{Evaluation Results}

\paragraph{Indoor-6} \Cref{tab:indoor6} shows results on Indoor-6. Our method performs on par with GLACE, and outperforms the baselines in the 5 minute SCR group.
Comparing ACE, DINO-ACE, and \methodname{} it can be seen that the DINO features improve the performance significantly over the ACE features, and a further improvement across metrics is achieved thanks to \methodname{}'s query-optimized, scene-agnostic coordinate regressor.
Comparing the scene coordinates predicted by ACE and \methodname{} in \cref{fig:scenecoords} indicates that ACE fails to cover some parts of the scene, suggesting that some features become too ambiguous to handle the multi-room scenes in Indoor-6.
GLACE, which addresses this issue through an added global encoding, performs similarly to our method.

\begin{figure*}[tb!]
    \centering
    \begin{subfigure}{0.49\linewidth}
        \centering
        \begin{tikzpicture}
            \scriptsize
            \node[inner sep=0pt, outer sep=0pt, anchor=north west] (ace) at (0,0) {\includegraphics[width=0.49\linewidth]{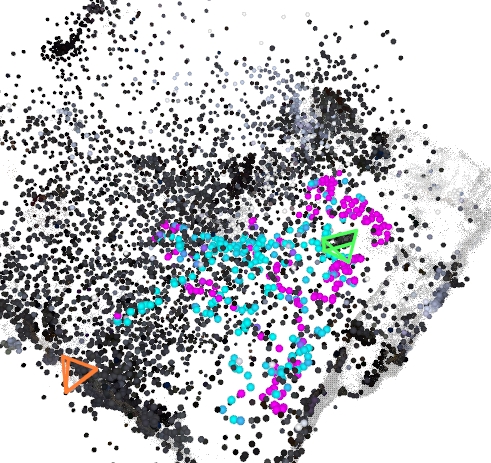}};
            \node[inner sep=0pt, outer sep=0pt, anchor=north west] (sv) at (0,0) {\includegraphics[width=0.25\linewidth]{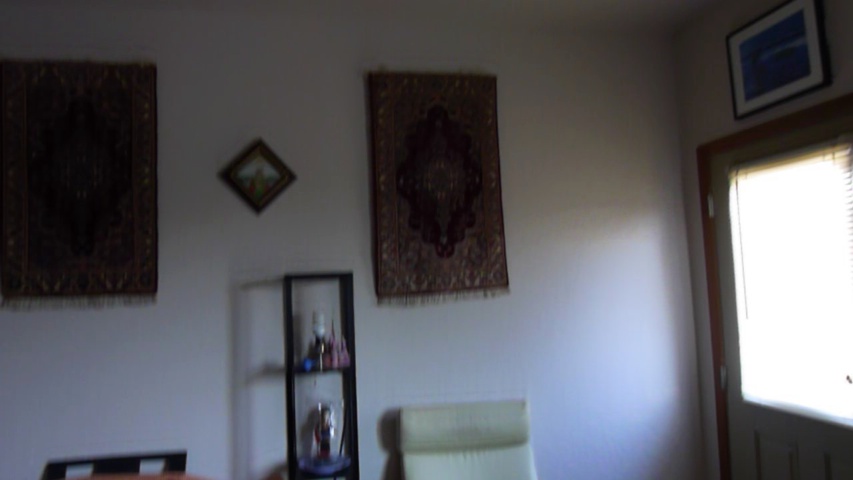}};
            \node[inner sep=2pt, outer sep=0pt, anchor=north west, fill=white, fill opacity=0.9, text opacity=1.0] (svlabel) at (sv.north west) {Mapping};
            \node[inner sep=2pt, outer sep=3pt, anchor=south, fill=white, fill opacity=0.9, text opacity=1.0] (methodlabel) at (ace.south) {ACE};
        \end{tikzpicture}%
        \begin{tikzpicture}[remember picture]
            \scriptsize
            \node[inner sep=0pt, outer sep=0pt, anchor=north east] (oursl) at (0,0) {\includegraphics[width=0.49\linewidth]{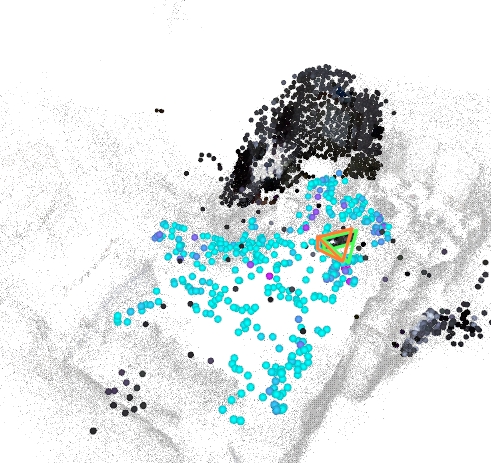}};
            \node[inner sep=0pt, outer sep=0pt, anchor=north west] (sv) at (oursl.north west) {\includegraphics[width=0.25\linewidth]{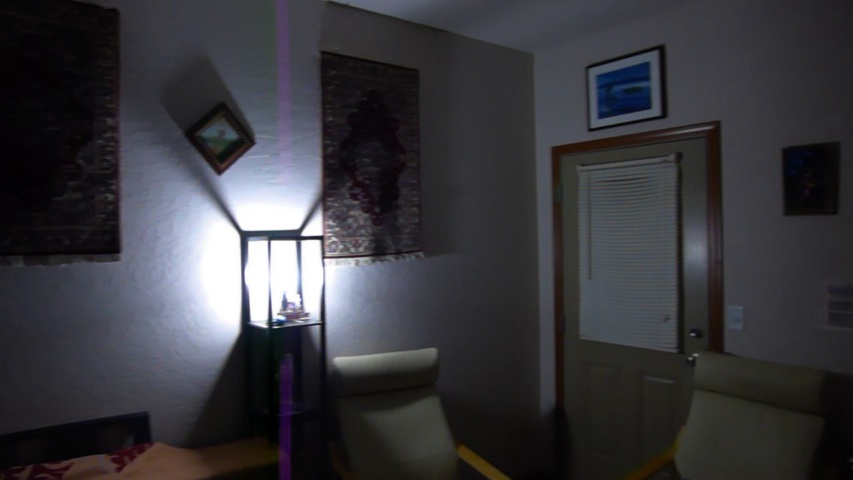}};
            \node[inner sep=2pt, outer sep=0pt, anchor=north west, fill=white, fill opacity=0.9, text opacity=1.0] (svlabel) at (sv.north west) {Query};
            \node[inner sep=2pt, outer sep=3pt, anchor=south, fill=white, fill opacity=0.9, text opacity=1.0] (methodlabel) at (oursl.south) {\methodname{}};
        \end{tikzpicture}%
        \caption{Indoor-6, Scene 5}
    \end{subfigure}
    \begin{subfigure}{0.49\linewidth}
        \centering
        \begin{tikzpicture}[remember picture]
            \scriptsize
            \node[inner sep=0pt, outer sep=0pt, anchor=north west] (acel) at (0,0) {\includegraphics[width=0.49\linewidth]{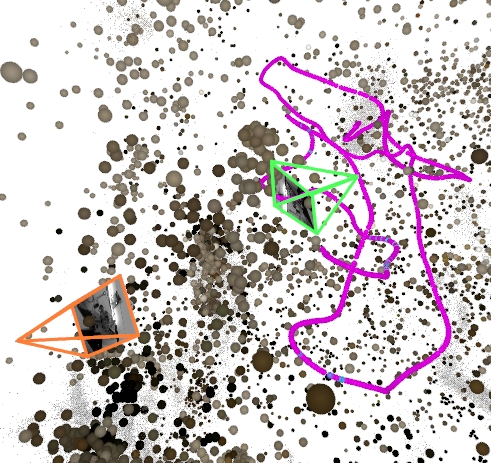}};
            \node[inner sep=0pt, outer sep=0pt, anchor=north west] (sv) at (0,0) {\includegraphics[height=0.25\linewidth]{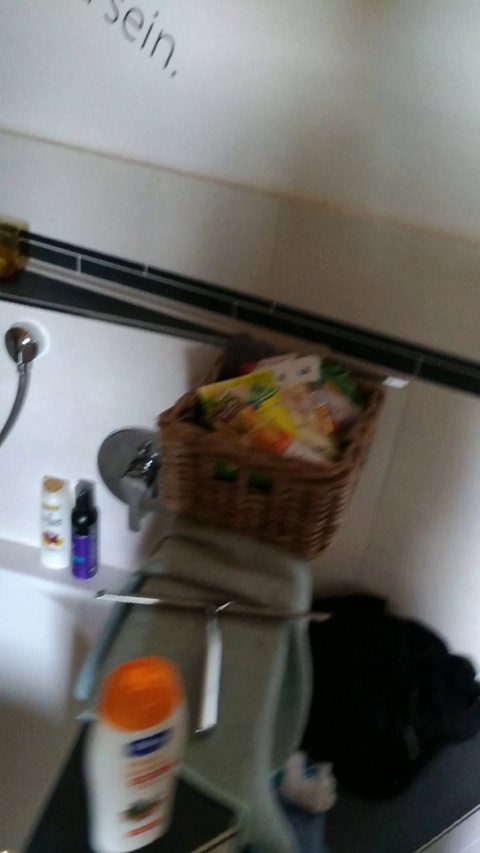}};
            \node[inner sep=2pt, outer sep=0pt, anchor=north west, fill=white, fill opacity=0.9, text opacity=1.0] (svlabel) at (sv.north west) {Mapping};
            \node[inner sep=2pt, outer sep=3pt, anchor=south, fill=white, fill opacity=0.9, text opacity=1.0] (methodlabel) at (acel.south) {ACE};
        \end{tikzpicture}%
        \begin{tikzpicture}
            \scriptsize
            \node[inner sep=0pt, outer sep=0pt, anchor=north east] (ours) at (0,0) {\includegraphics[width=0.49\linewidth]{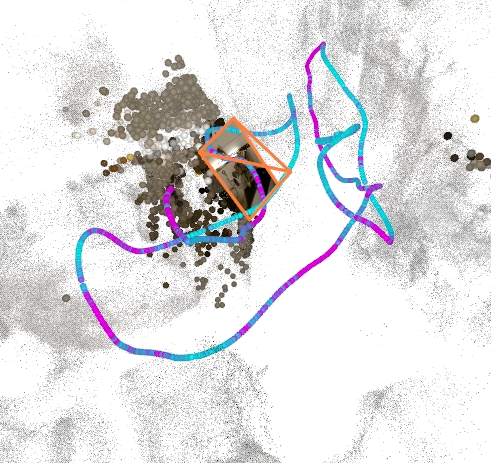}};
            \node[inner sep=0pt, outer sep=0pt, anchor=south east] (sv) at (ours.south east) {\includegraphics[height=0.25\linewidth]{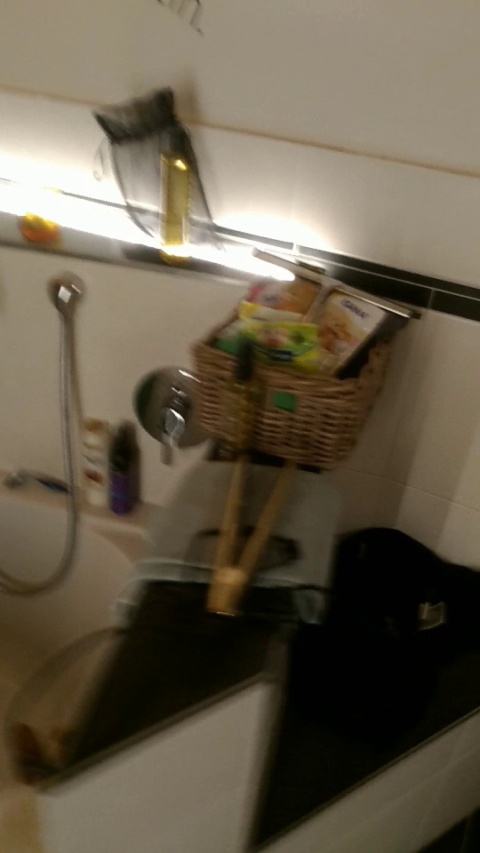}};
            \node[inner sep=2pt, outer sep=0pt, anchor=south east, fill=white, fill opacity=0.9, text opacity=1.0] (svlabel) at (sv.south east) {Query};
            \node[inner sep=2pt, outer sep=3pt, anchor=south, fill=white, fill opacity=0.9, text opacity=1.0] (methodlabel) at (ours.south) {\methodname{}};
        \end{tikzpicture}%
        \caption{RIO10, Scene 5}
    \end{subfigure}
    \begin{tikzpicture}[remember picture]
        \pgfplotsset{colormap={cool}{rgb(0)=(0,1,1); rgb(1000)=(1,0,1)}}
        \scriptsize
        \coordinate (center) at ($(oursl.south east)!0.5!(acel.south west)$);
        \shade[index of colormap=0 of cool,left color=.,color of colormap=1000 of cool,right color=.,local bounding box=colorbar] (center) ++ (-2.5, -0.35) rectangle ++(5,0.2);
        \node[anchor=north west, inner xsep=0] (minlabel) at (colorbar.south west) {5\textdegree\,/ $5\,\mathrm{cm}$\strut};
        \node[anchor=north east, inner xsep=0] (maxlabel) at (colorbar.south east) {50\textdegree\,/ $50\,\mathrm{cm}$\strut};
        \node[anchor=north] (midlabel) at (colorbar.south) {$\mathrm{max}(\Delta\text{\textdegree}, \Delta\,\mathrm{cm})$\strut};
        \pgfresetboundingbox
        \path[use as bounding box] (0,0);
    \end{tikzpicture}%
    \caption{\textbf{Qualitative comparison for ACE and \methodname{}.} The full ground-truth trajectory is shown as dots with color indicating the error. The estimated scene coordinates for one query image are shown along with the \textcolor{orange}{estimated} and \textcolor{green!80!black}{ground-truth} camera pose (mapping image shows a close mapping image chosen based on manual inspection). ACE fails to generalize to changing lighting conditions (a) and changing objects (b), while \methodname{}'s predictions remain sensible resulting in a correct pose estimate (best viewed digitally).}
    \label{fig:qualitativeresults}
    \vspace*{-\baselineskip}
\end{figure*}
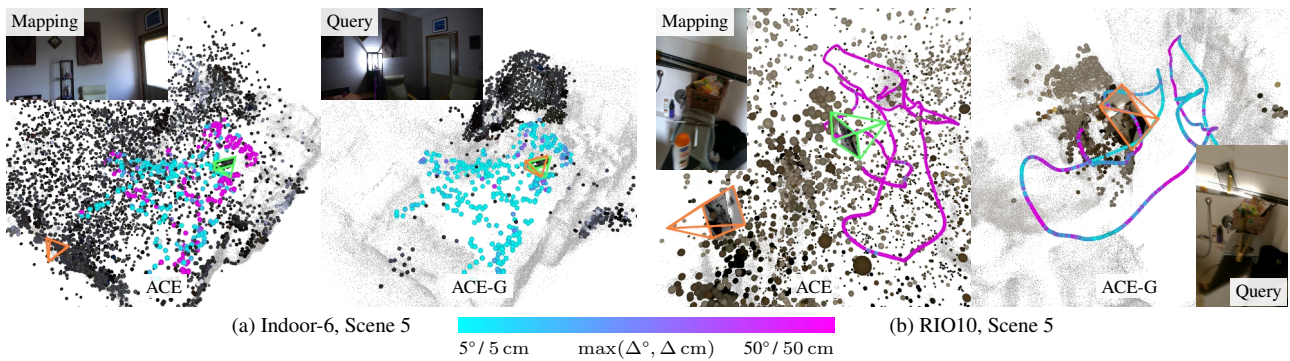

\paragraph{RIO10} \Cref{tab:rio10} shows a significant improvement in terms of robustness to changing object arrangements, compared to other SCR methods
(\eg, note that for all scenes the median error is consistently lower than $<1\,\mathrm{m}$). MASt3R fails to compute a pose for more than 50\% of images for some of the scenes (marked as N/A) highlighting the challenging nature of this dataset. Reloc3r, the only method achieving results similar to ours, requires access to the mapping images to perform relocalization.
\Cref{fig:qualitativeresults} further highlights \methodname{}'s improved generalization ability. %

\paragraph{Cambridge Landmarks}  \Cref{tab:cambridge} reports median errors for Cambridge Landmarks.
The results show that among the SCR methods, GLACE, which aims for scalability to larger scenes performs best.
Our method performs slightly worse than ACE on average.
One reason for the different trend of results compared to Indoor-6 and RIO10 is the relatively small difference between mapping and query present in this dataset.
In addition, note that our pre-training did not include scenes comparable in size and content to the ones present in this dataset.
Interestingly, our method still outperforms DINO-ACE overall, that is, our pre-training still improved performance.

\begin{table}[tb!]
    \centering
    \caption{\textbf{Results on Cambridge Landmarks.} Each cell contains in order: median translation error in $\mathrm{cm}$ and median rotation error in degrees. Best in each SCR group \textbf{highlighted}. Despite this dataset being out-of-distribution with respect to our training datasets, \methodname{} achieves competitive results. Our architecture still improves notably over the DINO-ACE baseline.}%
    \vspace{-0.5\baselineskip}%
    \scriptsize%
    \renewcommand{\arraystretch}{1.1}%
    \setlength{\tabcolsep}{2pt}%
    \begin{NiceTabular}{@{}lcccccc@{}}
    \toprule
       \quad$(\mathrm{cm}\,/\,\si{\degree})$ &  GC & KC & OH & SF & SMC & Avg. \\
    \midrule
        Reloc3R \cite{dong2025reloc3r} & 122 / 0.7 & 42 / 0.4 & 62 / 0.6 & 13 / 0.6 & 34 / 0.6 & 55 / 0.6 \\
        MASt3R+K. \cite{leroy2024grounding} & 12.6 / 0.6 & 7.3 / 0.1  & 15.2 / 0.3 / & 3.8 / 0.2 & 4.1 / 0.1 & 9.1 / 0.2 \\
    \midrule
        GLACE & \textbf{18.0} / \textbf{0.1} & \textbf{18.0} / \textbf{0.3} & \textbf{17.9} / \textbf{0.4} & \textbf{4.5} / \textbf{0.2} & \textbf{8.4} / \textbf{0.3} & \textbf{13.5} / \textbf{0.3} \\
        Ours (25 min.) & 50.6 / 0.3 & 21.1 / 0.3 & 22.5 / 0.5 & 7.4 / 0.3 & 24.9 / 0.8 & 25.3 / 0.4 \\[\aboverulesep]
\dotrule \noalign{\vskip \belowrulesep}
        ACE & \textbf{42.2} / \textbf{0.2} & 26.7 / \textbf{0.4} & 30.7 / 0.6 & \textbf{5.3} / 0.3 & \textbf{20.6} / \textbf{0.6} & \textbf{25.1} / \textbf{0.4} \\
        DINO-ACE & 61.4 / 0.3 & 38.9 / 0.4 & 39.8 / 0.6 & 5.3 / \textbf{0.3} & 74.0 / 1.8 & 43.9 / 0.7  \\
        Ours (5 min.) & 68.2 / 0.4 & \textbf{26.3} / 0.4 & \textbf{28.7} / \textbf{0.5} & 7.8 / 0.3 & 44.2 / 1.4 & 35.1 / 0.6 \\
    \bottomrule
    \end{NiceTabular}%
    \label{tab:cambridge}%
    \vspace*{-\baselineskip}%
\end{table}

\begin{table}[tb!]
    \centering
    \caption{\textbf{Accuracy under fine thresholds.} Proportion of frames correctly localized under (5\textdegree, $5\,\mathrm{cm}$) error threshold for Indoor-6 and RIO10, and (10\textdegree, $10\,\mathrm{cm}$) error for Cambridge. Best in each SCR group \textbf{highlighted}.}%
    \vspace{-0.5\baselineskip}%
    \scriptsize%
    \begin{NiceTabular}{@{}lccc@{}}
    \toprule
     \quad ($\%$) & Indoor-6 & RIO10 & Cambridge \\
    \midrule
        Reloc3R \cite{dong2025reloc3r} & 37.8 & 9.7 & 11.8 \\
        MASt3R+Kapture \cite{leroy2024grounding} & 77.0 & 13.0 & 65.0 \\
    \midrule
        GLACE & \bfseries 69.6 & \bfseries 9.8 & \bfseries 42.5 \\
        Ours (25 min.) & 64.7 & 7.4 &  24.6\\[\aboverulesep]
\dotrule \noalign{\vskip \belowrulesep}
        ACE & 33.5 & 4.2 & \bfseries 26.3 \\
        DINO-ACE & 44.4	& 2.9 & 17.3 \\
        Ours (5 min.) & \bfseries 54.7 & \bfseries 5.7 & 17.9 \\
    \bottomrule
    \end{NiceTabular}%
    \label{tab:finer}%
    \vspace*{-1\baselineskip}%
\end{table}

\paragraph{Fine Accuracy} \Cref{tab:finer} reports results for finer accuracy thresholds across datasets.
Compared to the previously reported, more lenient thresholds and mean median metrics, \methodname{} performs slightly worse than GLACE (compare relative performance to GLACE in \cref{tab:indoor6} and \cref{tab:rio10}). We hypothesize that the slightly worse performance for fine accuracies can be attributed to the lower resolution of DINO features -- patch size of 14 pixels across vs. 8 pixels for ACE and GLACE's convolutional image encoder -- resulting in fewer estimated correspondences per image and possibly less accurate geometry estimation. Nevertheless, for indoor datasets \methodname{} remains better than ACE and DINO-ACE. Notably, on RIO10, DINO-ACE performs worse than ACE, but \methodname{} performs better than both, highlighting the advantage of the pre-trained coordinate regressor.

\subsection{Analysis}

\begin{figure}[tb]
    \centering
    \scriptsize
    \renewcommand{\arraystretch}{1.1}%
    \setlength{\tabcolsep}{2pt}%
    \begin{tabular}{rC{0.7cm}C{0.7cm}C{0.7cm}C{0.7cm}C{0.7cm}C{0.7cm}}
        \toprule
        \quad ($\%$) & SN & SN++ & ARK & MF & BMVG & WR \\
        \midrule
        Mapping-only & 48.2 & 39.8 & 39.4 & 38.4 & 28.9 & 26.3 \\
        Mapping+query & 56.2 & 49.2 & 49.3 & 42.3 & 32.7 & 32.7  \\
        Change & \cellcolor{green!30}+8.0 & \cellcolor{green!30}+9.4 & \cellcolor{green!30}+9.9 & \cellcolor{green!30}+3.9 & \cellcolor{green!30}+3.8 & \cellcolor{green!30}+6.4\\
        \bottomrule
    \end{tabular}\\
    \includegraphics{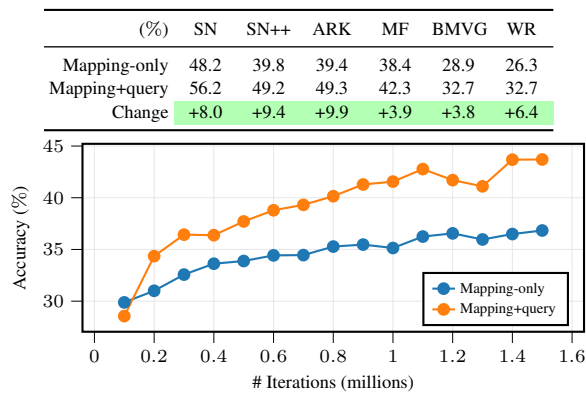}
    \caption{\textbf{Query training.} Top: mean accuracy at the $(20\si{\degree},20\,\mathrm{cm})$ error threshold after 1.5M iterations, and relative change attributable to query training.
    Bottom: mean accuracy on the validation splits as training progresses.}
    \label{fig:querytrain}
    \vspace*{-\baselineskip}
\end{figure}

\paragraph{Query Training} In \cref{fig:querytrain} we show the performance on the validation sets with and without query iterations during the pre-training of the scene-agnostic coordinate regressor.
The ``mapping+query'' model alternates between mapping and query optimization iterations, whereas the ``mapping'' model only performs mapping iterations (\cf \cref{sec:multiscene}).
By including the query training optimization steps we can see a clear improvement across dataset validation performance, increasing the robustness and generalization of \methodname{}. %

\begin{figure}[tb]
    \centering
    \centering
    \begin{tikzpicture}
        \scriptsize
        \node[inner sep=0pt, outer sep=0pt, anchor=north west] (ace) at (0,0) {\includegraphics[width=0.49\linewidth]{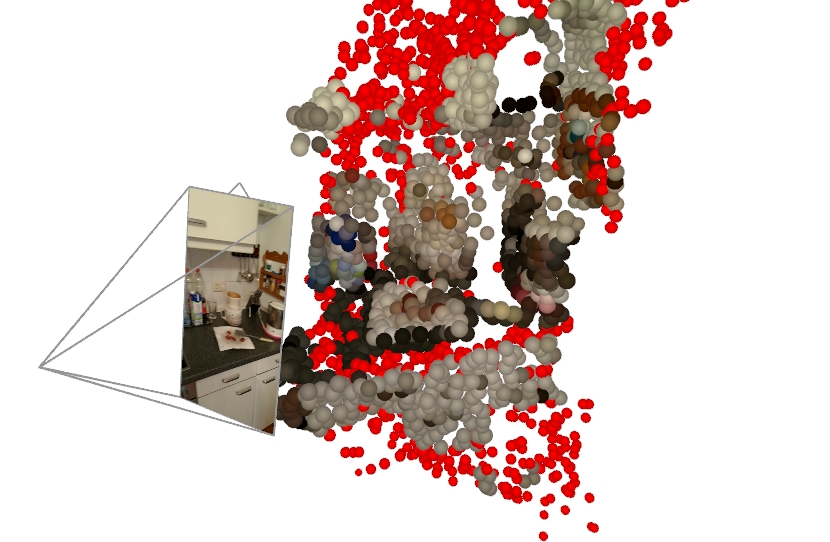}};
        \node[inner sep=0pt, outer sep=0pt, anchor=south east] (sv) at (ace.south east) {\includegraphics[width=0.14\linewidth]{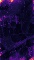}};
    \end{tikzpicture}%
    \begin{tikzpicture}
        \scriptsize
        \node[inner sep=0pt, outer sep=0pt, anchor=north west] (ace) at (0,0) {\includegraphics[width=0.49\linewidth]{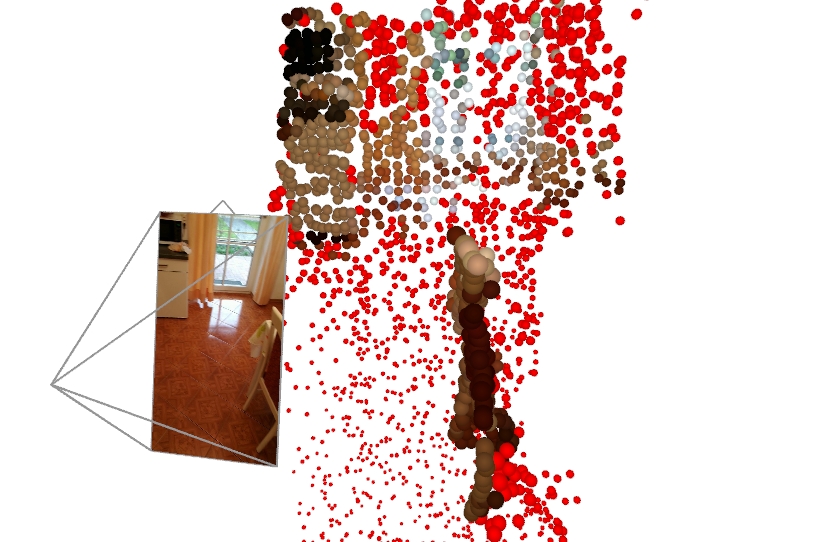}};
        \node[inner sep=0pt, outer sep=0pt, anchor=south east] (sv) at (ace.south east) {\includegraphics[width=0.14\linewidth]{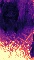}};
    \end{tikzpicture}%
    \vspace{1mm}
    \scriptsize
    \renewcommand{\arraystretch}{1.1}%
    \setlength{\tabcolsep}{2pt}%
    \begin{tabular}{rC{0.7cm}C{0.7cm}C{0.7cm}C{0.7cm}C{0.7cm}C{0.7cm}}
        \toprule
        \quad ($\%$) & SN & SN++ & ARK & MF & BMVG & WR \\
        \midrule
        No prefiltering & 73.2 & 65.6 & 63.4 & 57.3 & 56.6 & 42.8 \\
        With prefiltering & 77.0 & 70.3 & 65.3 & 63.5 & 57.2 & 51.8  \\
        Change & \cellcolor{green!30}+3.8&\cellcolor{green!30}+4.7&\cellcolor{green!30}+1.9&\cellcolor{green!30}+6.2&\cellcolor{green!30}+0.6 &\cellcolor{green!30}+9.0 \\
        \bottomrule
    \end{tabular}%
    \caption{\textbf{Uncertainty prefiltering.} Quantile-based filtering removes points (\textcolor{red}{highlighted}) based on the difficulty of the query image, improving results.}%
    \label{fig:prefiltering}%
    \vspace*{-2\baselineskip}%
\end{figure}

\paragraph{Uncertainty-Based Prefiltering} As described in \cref{sec:relocalization}, we use the uncertainty predicted by the model to prefilter the scene coordinates used for registration.
\Cref{fig:prefiltering} shows the filtered and raw coordinates for two examples. The adaptive threshold adjusts the number of estimated correspondences used in PnP based on the image difficulty. For a query image with small mapping-query gap and large visual overlap, most estimates will have similar uncertainty and will pass the quantile-based filter. For a query image with larger mapping-query gap the uncertainty varies depending on image content and previously seen mapping data and the filter will remove the high uncertainty estimates.

\section{Conclusion}
We presented \methodname{}, a transformer-based approach to scene coordinate regression explicitly optimized to generalize to novel views and changing conditions. Built on top of DINO features, our approach demonstrates improved invariance to changing lighting conditions and robustness to environment changes. %
It outperforms state-of-the-art scene coordinate regression methods on difficult datasets and achieves competitive performance on 
large-scale datasets without major mapping-query differences.
Overall, \methodname{} is a strong alternative to traditional SCRs that are optimized per-scene, and an important conceptual step towards enabling further learning-based approaches.

\clearpage

\FloatBarrier
{
    \small
    \bibliographystyle{ieeenat_fullname}
    \bibliography{main}
}

\clearpage
\setcounter{page}{1}
\maketitlesupplementary

\section{Hyperparameters}\label{sec:implementationdetails}

\paragraph{Architecture} 
We use DINOv2-L with registers \cite{oquab2024dinov,darcet2024vision} as our image encoder and use $N=12$ cross-attention-only transformer blocks (\cf \cref{fig:network}). 

\paragraph{Pre-Training}
We train our network for 4.4M map code iterations (resulting in 440\,000 mapping and 440\,000 query iterations for the head, because we are only updating the head in every 10th iteration, \cf \cref{sec:multiscene}) on 8 A100 GPUs with a scene batch size $N_\mathrm{spb}=200$ and patch batch size $N_\mathrm{pps}=512$. Each map code is optimized between 6000 and 10000 iterations with $N_\mathrm{qstandby}=5000$. To focus optimization on solvable patches we found it beneficial to only use the lowest 30\% losses in each batch. We use AdamW for map codes and network weights with a learning rate of 0.0001 without learning rate scheduler. During pre-training we use a map code size of $N_\mathcal{C}=1024$.

\paragraph{Novel Scene Mapping}
We use slightly varying parameters for our 5 minute and 25 minute configuration following manual tuning on validation scenes. In the 5 minute setup a maximum buffer size of 4M patches, 1000 iterations, and a batch size of 40960 is used. In the 25 minute setup we spend more time budget on the buffer creation using 8M patches, increase the number of iterations to 4000, and increase the batch size to 51200.
In both cases, AdamW with a one cycle learning rate schedule with maximum learning rate 0.002 is used. We use $N_\mathcal{C}=4096$ resulting in 12 MB maps (full precision). During optimization we apply dropout on the image features with a dropout probability of $10\%$.

\section{Datasets}
Every combination of mapping images yields a unique map code after optimization and every other image in a sequence can potentially aid in improving the generalization performance of the coordinate regressor. Therefore, we randomly generate multiple mapping-query configurations per scene taking into account specific dataset characteristics. For most datasets, sequences are ordered in time which gives a strong clue for which images are likely covisible. Therefore, we follow an interval-based configuration scheme where the sorted image sequence is split into disjoint subsets serving as the mapping and query portion. For unsorted datasets we adjust parameters such that empirically most query views should still be solvable while also including challenging views with little visual overlap.

We follow two sampling schemes: an interspersed one, in which mapping and query intervals of varying length alternate throughout the sequence; and a query-mapping-query scheme, where a mapping interval of varying length is surrounded by two query intervals of varying length.

For ARKitScenes and ScanNet++, we first sample a mapping interval, then find covisible image pairs given the image pair information published by \cite{wang2024dust3r} and sample a short interval around the image known to be covisible.

Beyond this interval-based sampling, we randomly switch mapping and query sequences for the MapFree dataset and randomly mirror scenes. Finally, a random rotation is applied every time a new mapping-query configuration is added to the active scenes.

\begin{figure*}[tbhp]
    \centering
    \begin{subfigure}{0.49\linewidth}
        \begin{tikzpicture}
            \scriptsize
            \node[inner sep=0pt, outer sep=0pt, anchor=north west] (dataset) at (0,0) {\includegraphics[width=\linewidth]{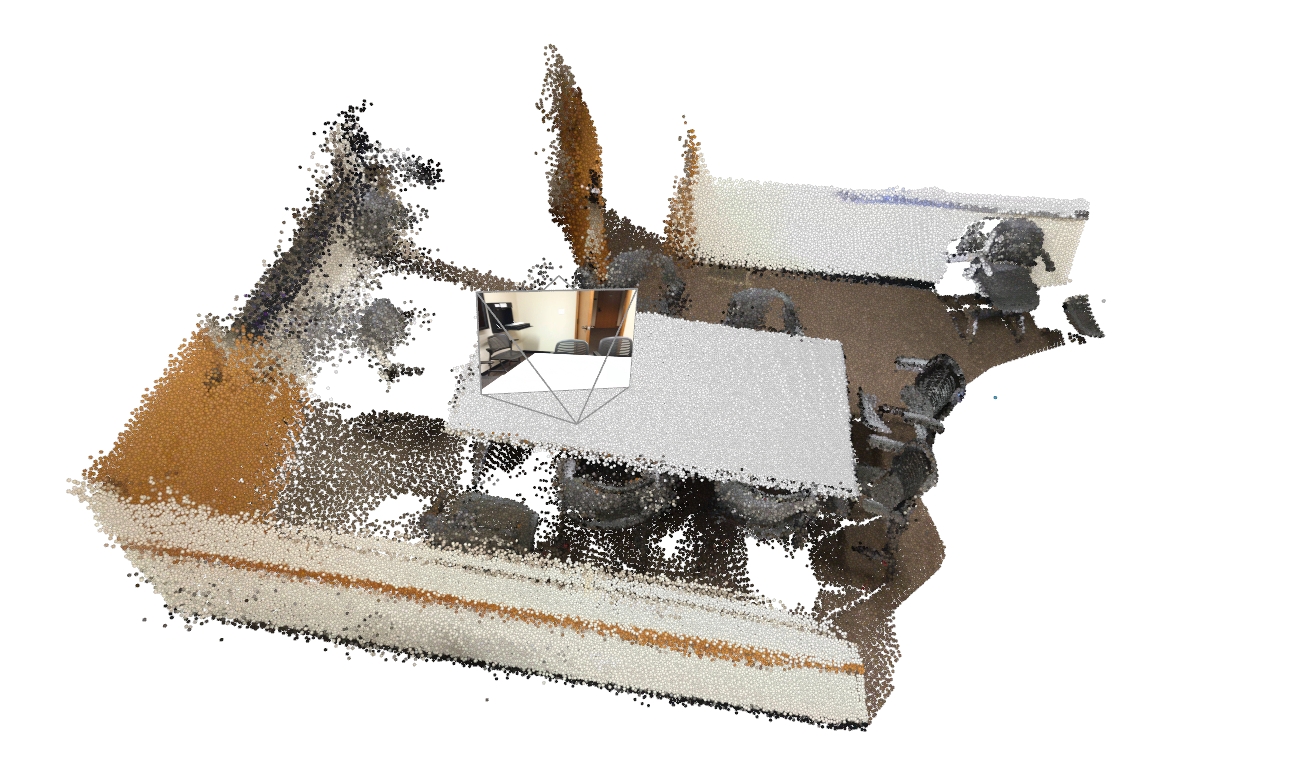}};
            \node[inner sep=0pt, outer sep=0pt, anchor=south east, draw] (query) at (dataset.south east) {\includegraphics[width=0.35\linewidth]{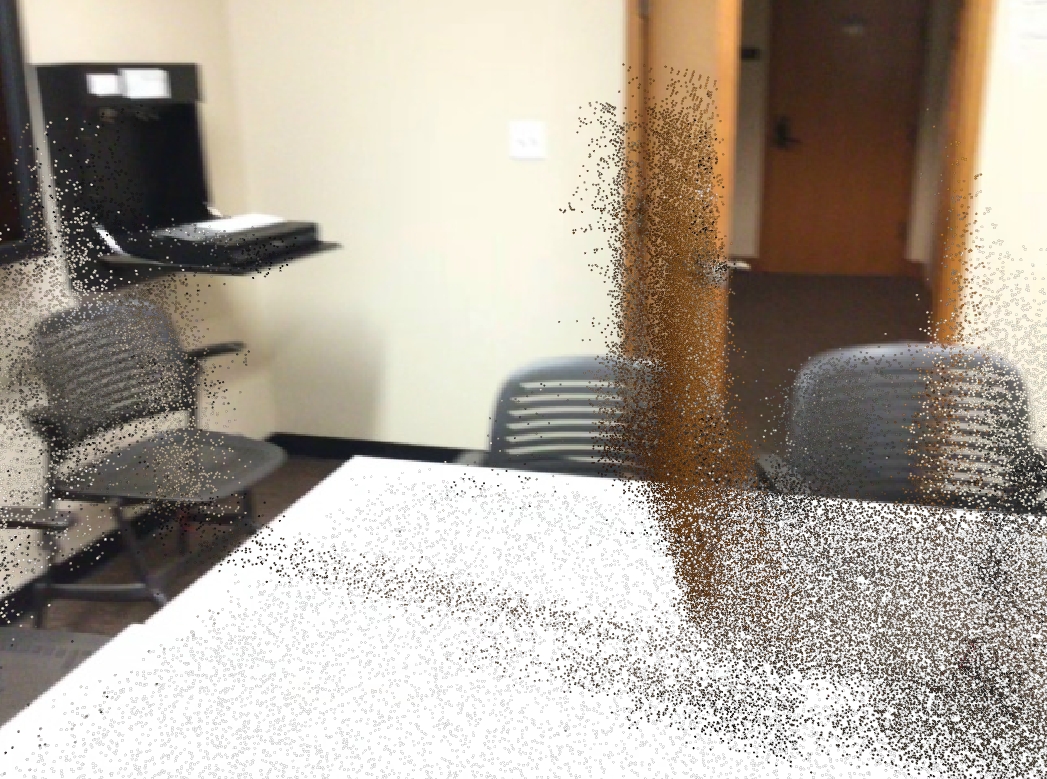}};
        \end{tikzpicture}%
        \subcaption{ScanNet \cite{dai2017scannet}}
    \end{subfigure}
    \begin{subfigure}{0.49\linewidth}
        \begin{tikzpicture}
            \scriptsize
            \node[inner sep=0pt, outer sep=0pt, anchor=north west] (dataset) at (0,0) {\includegraphics[width=\linewidth]{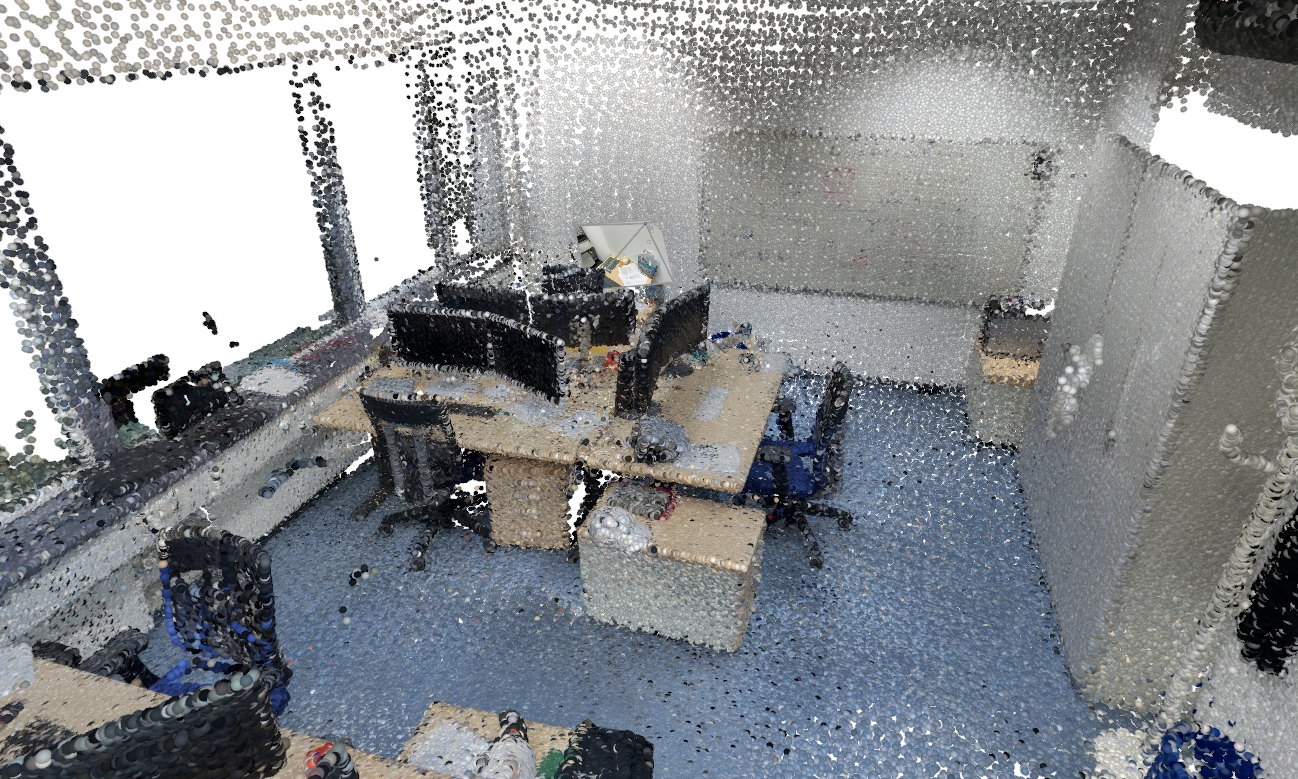}};
            \node[inner sep=0pt, outer sep=0pt, anchor=south east, draw] (query) at (dataset.south east) {\includegraphics[width=0.35\linewidth]{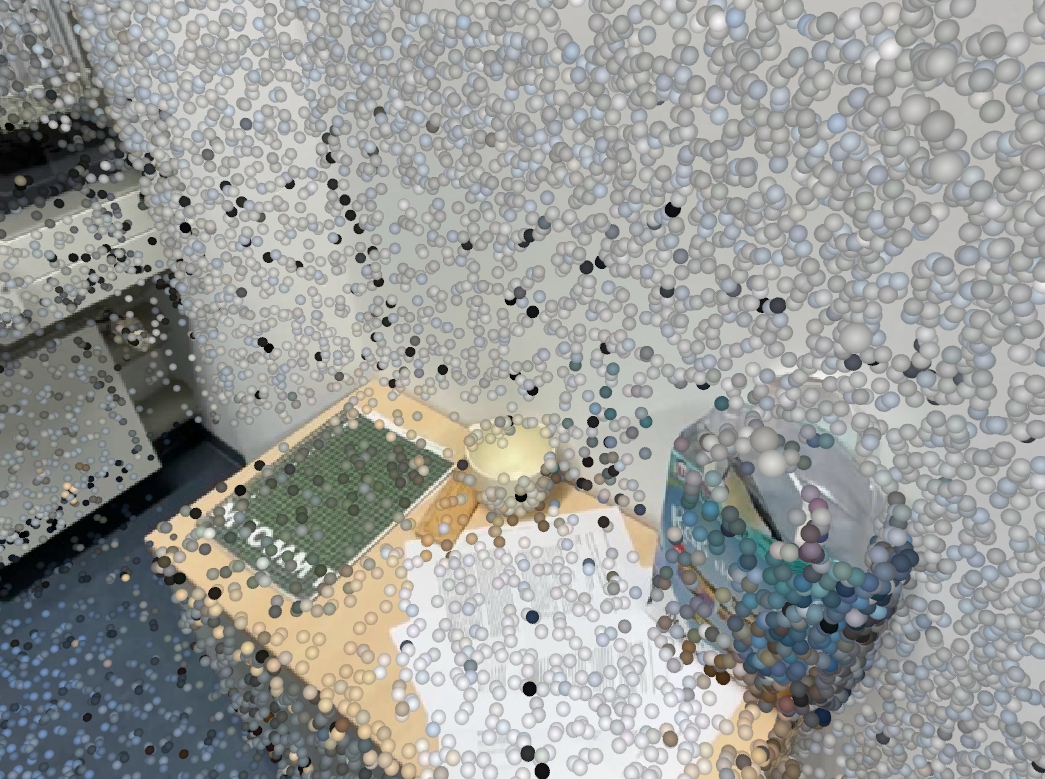}};
        \end{tikzpicture}%
        \subcaption{ScanNet++ \cite{yeshwanth2023scannet++}}
    \end{subfigure}
    \begin{subfigure}{0.49\linewidth}
        \begin{tikzpicture}
            \scriptsize
            \node[inner sep=0pt, outer sep=0pt, anchor=north west] (dataset) at (0,0) {\includegraphics[width=\linewidth]{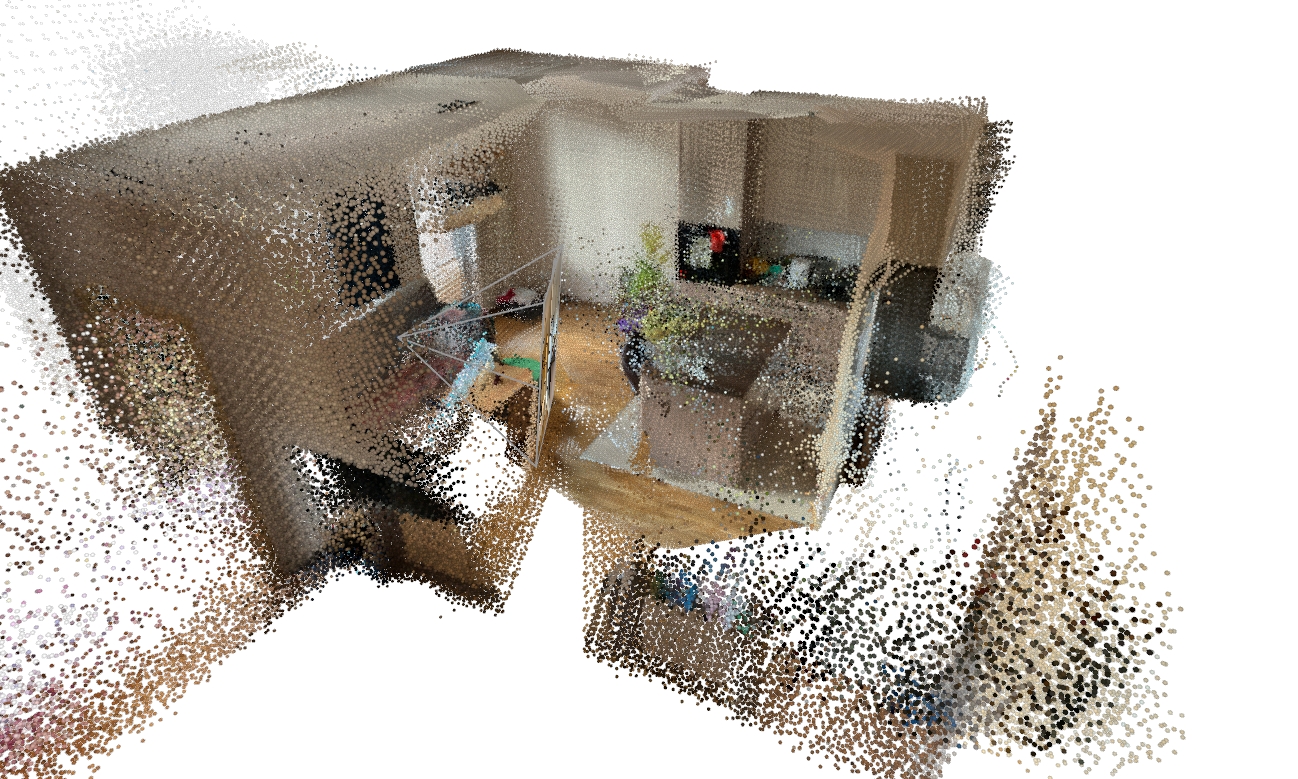}};
            \node[inner sep=0pt, outer sep=0pt, anchor=south east, draw] (query) at (dataset.south east) {\includegraphics[height=0.35\linewidth]{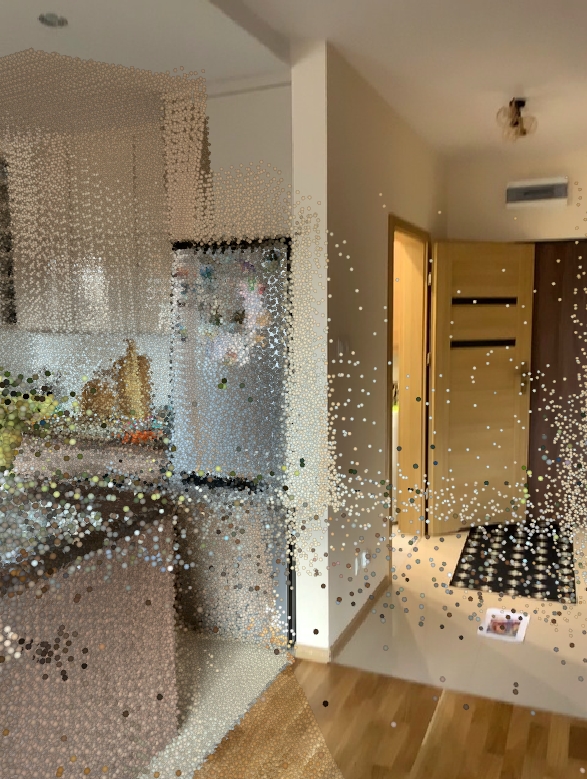}};
        \end{tikzpicture}%
        \subcaption{ARKitScenes \cite{baruch2021arkitscenes}}
    \end{subfigure}
    \begin{subfigure}{0.49\linewidth}
        \begin{tikzpicture}
            \scriptsize
            \node[inner sep=0pt, outer sep=0pt, anchor=north west] (dataset) at (0,0) {\includegraphics[width=\linewidth]{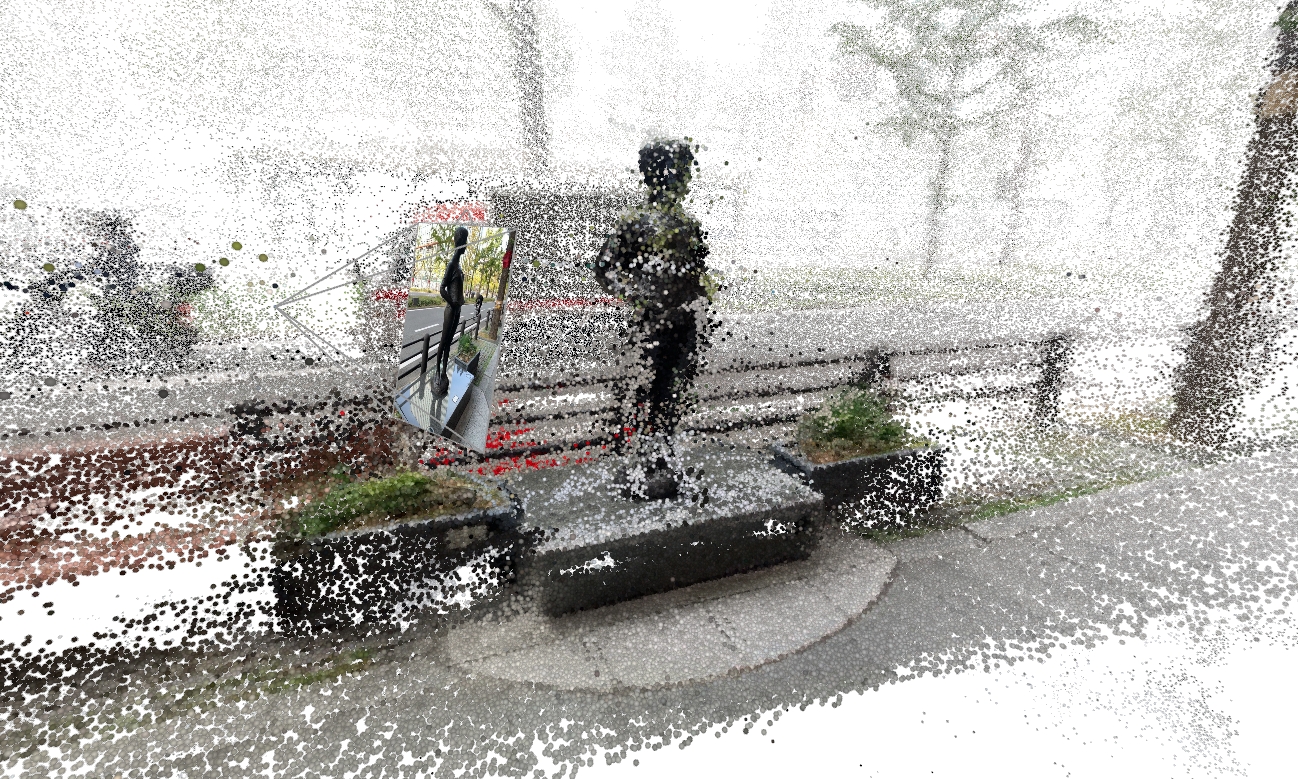}};
            \node[inner sep=0pt, outer sep=0pt, anchor=south east, draw] (query) at (dataset.south east) {\includegraphics[height=0.35\linewidth]{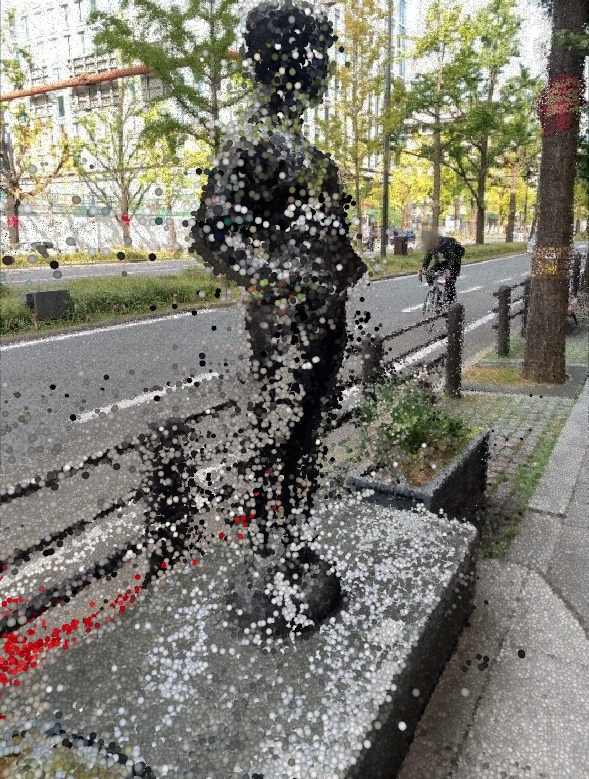}};
        \end{tikzpicture}%
        \subcaption{MapFree \cite{arnold2022map}}
    \end{subfigure}
    \begin{subfigure}{0.49\linewidth}
        \begin{tikzpicture}
            \scriptsize
            \node[inner sep=0pt, outer sep=0pt, anchor=north west] (dataset) at (0,0) {\includegraphics[width=\linewidth]{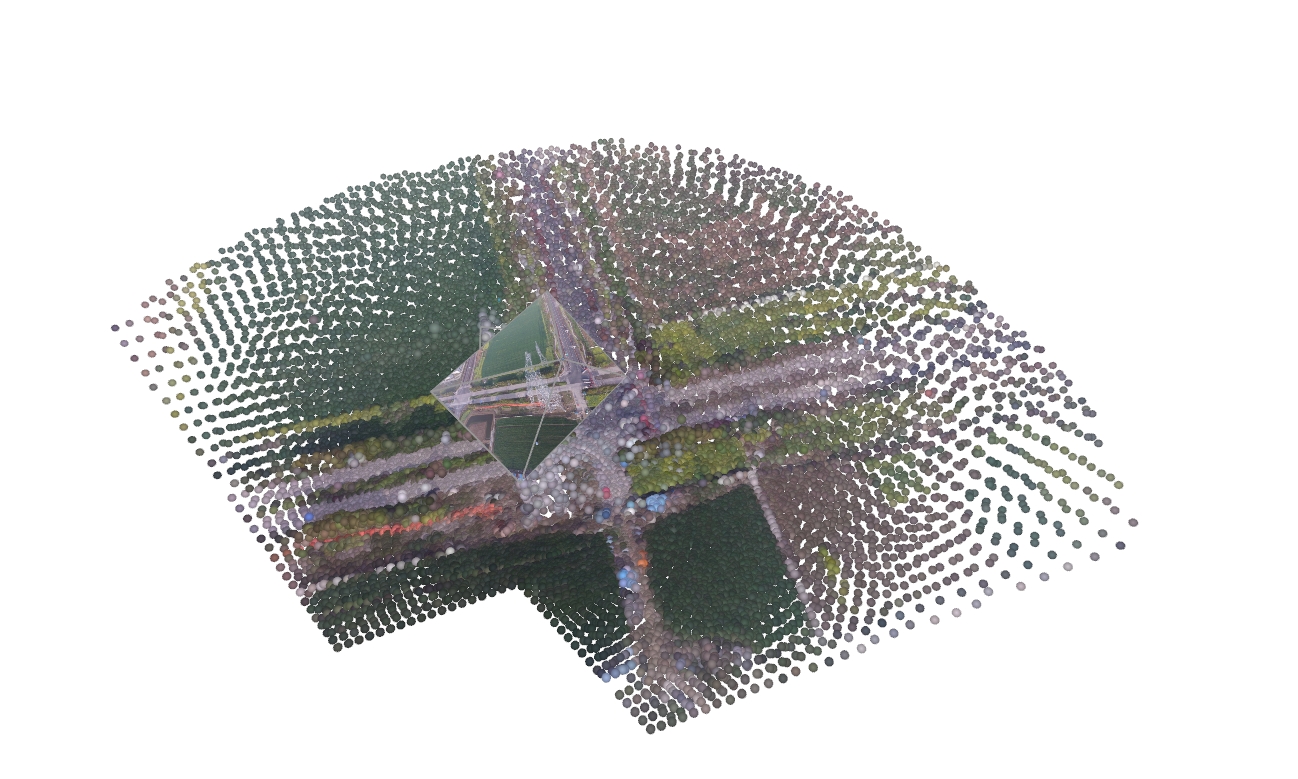}};
            \node[inner sep=0pt, outer sep=0pt, anchor=south east, draw] (query) at (dataset.south east) {\includegraphics[width=0.35\linewidth]{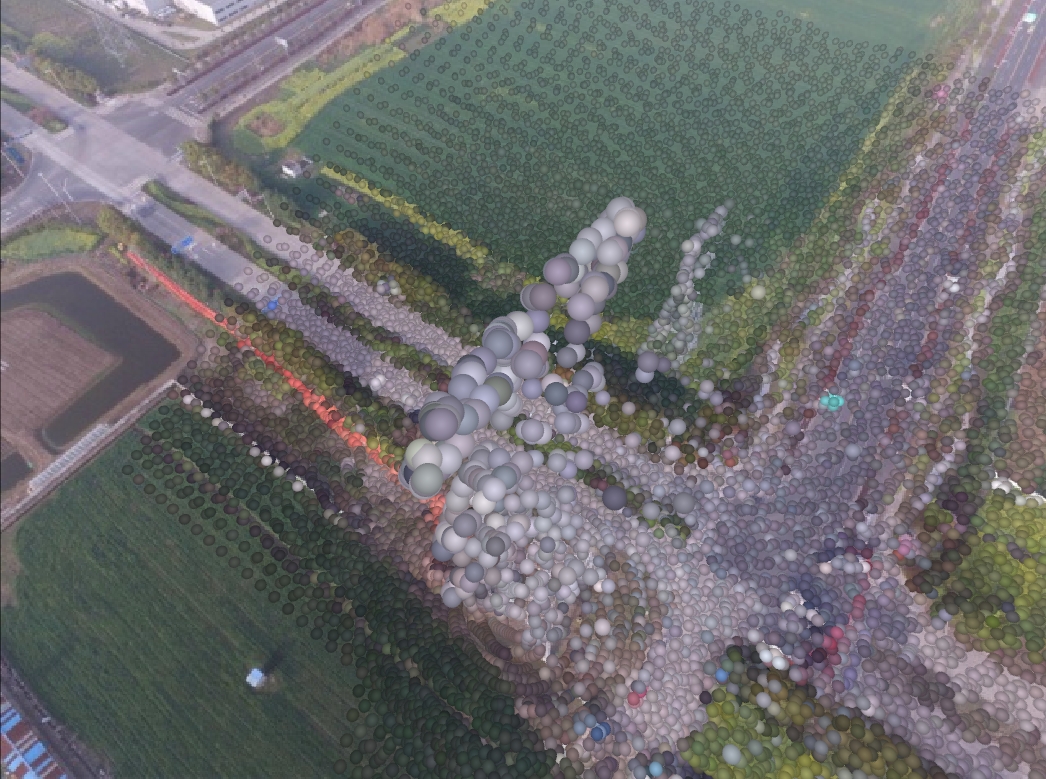}};
        \end{tikzpicture}%
        \subcaption{BlendedMVG \cite{yao2020blendedmvs}}
    \end{subfigure}
    \begin{subfigure}{0.49\linewidth}
        \begin{tikzpicture}
            \scriptsize
            \node[inner sep=0pt, outer sep=0pt, anchor=north west] (dataset) at (0,0) {\includegraphics[width=\linewidth]{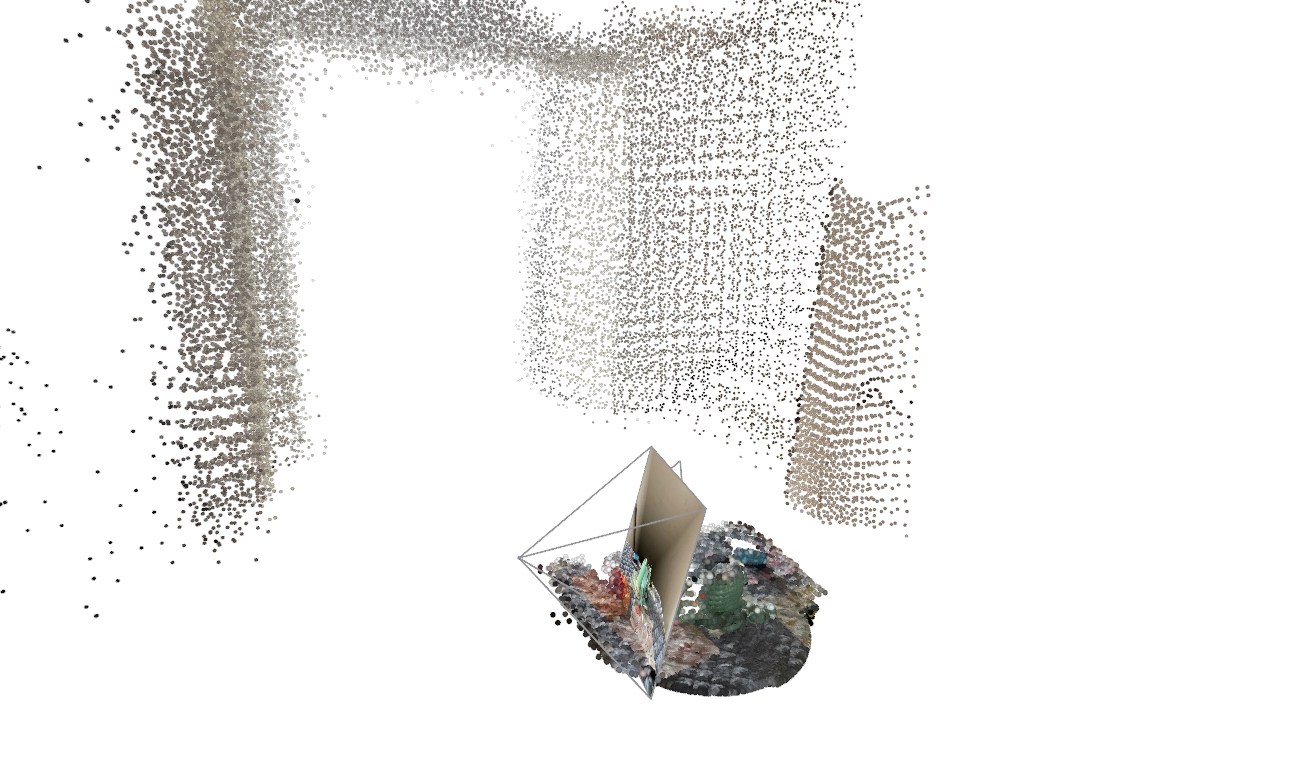}};
            \node[inner sep=0pt, outer sep=0pt, anchor=south east, draw] (query) at (dataset.south east) {\includegraphics[height=0.35\linewidth]{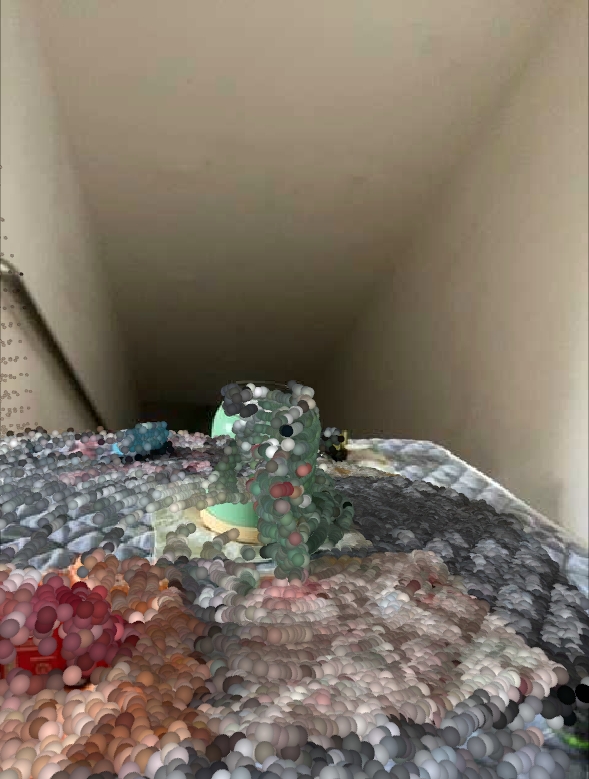}};
        \end{tikzpicture}%
        \subcaption{WildRGBD \cite{xia2024rgbd}}
    \end{subfigure}
    \caption{\textbf{Mapping-query splits used for training.} The 3D points show the accumulated scene coordinates from all mapping views. The inset shows projected ground-truth points in a query view. Note that overlap between mapping points and query view varies significantly.}
    \label{fig:datasetsamples}
\end{figure*}

\Cref{fig:datasetsamples} visualizes a mapping-query split for each included dataset.

\begin{table}[tb!]
    \caption{\textbf{Image encoder analysis}. Accuracy, in terms of median position error (in cm), on 7Scenes, 12Scenes, Indoor-6 and RIO10 (top) with per-scene results for 7Scenes (bottom) for different image encoders. \textbf{Best} and {\ul second best} highlighted.}\label{tab:encoder}
    \scriptsize
    \centering
    \renewcommand{\arraystretch}{1.1}%
    \setlength{\tabcolsep}{2pt}%
    \begin{tabular}{@{}lccccc@{}}
    \toprule
        & \multicolumn{2}{c}{Static} && \multicolumn{2}{c}{Dynamic}  \\ 
    \cmidrule(){2-3}\cmidrule(){5-6}
        & 7S           & 12S          && I6           & R10           \\ \midrule
    ACE w/ ACE enc.\ & \textbf{1.1} & \textbf{0.7} && 11.0         & 358.4         \\
    ACE w/ DINOv2 enc. (``DINO-ACE'')\ & 7.2          & 1.9          && {\ul 5.6}    & {\ul 83.8}    \\
    ACE-G w/ ACE enc.\ (Ours) & {\ul{1.3}} & \textbf{0.7} && 11.3 & 144.5 \\
    ACE-G w/ DINOv2 enc.\ (Ours) & \tikz[remember picture,baseline]{\node[anchor=base] (topanchor) {4.6}}    & {\ul 1.2}    && \textbf{4.5} & \textbf{41.1} \\ 
    \bottomrule
    \end{tabular}\\[0.5cm]
    \begin{tikzpicture}[overlay, remember picture]
        \draw (topanchor.south west) ++ (0, -0.18) -- ++(-4.9,-0.4);
        \draw (topanchor.south east) ++ (0, -0.18) -- ++(2.88,-0.4);
    \end{tikzpicture}
    \centering
    \scriptsize
    \renewcommand{\arraystretch}{1.1}%
    \setlength{\tabcolsep}{2pt}%
    \begin{tabular}{@{}lccccccc@{}}
    \toprule 
    & Chess & Fire & Heads & Off. & Pump. & RK & Stairs \\ 
    \midrule
    ACE w/ ACE enc.         & \textbf{0.6} & \textbf{0.8}  & \textbf{0.6}  & {\ul{1.1}}    & \textbf{1.2}  & \textbf{0.8}  & \textbf{2.8} \\
    ACE w/ DINOv2 enc. (``DINO-ACE'') & {\ul{0.9}}   & {\ul{1.4}}    & {\ul{0.8}}    & 1.4           & 1.8           & {\ul{1.1}}    & 43.9 \\
    ACE-G w/ ACE enc. (Ours) & \textbf{0.6} & \textbf{0.8}  & \textbf{0.6}  & \textbf{1.0}  & {\ul{1.4}}    & \textbf{0.8}  & {\ul{3.8}} \\
    ACE-G w/ DINOv2 enc. (Ours) & 1.0          & 1.5           & 0.9           & 1.4           & 1.8           & {\ul{1.1}}    & 24.5 \\ 
    \bottomrule
    \end{tabular}
\end{table}

\section{Additional Experiments}
\subsection{Image Encoder}
To better understand the interplay of our pre-training and the image encoder, we report additional results of ACE and \methodname{} paired with ACE's fully-convolutional image encoder and DINOv2 in \cref{tab:encoder}. In addition to the datasets reported in the main paper, we include 7Scenes \cite{shotton2013scene} and 12Scenes \cite{valentin2016learning}. Datasets can be grouped into \emph{static} (7Scenes, 12Scenes) and \emph{dynamic} (Indoor-6, RIO10), depending on whether there are environment and lighting changes between mapping and query images.

In summary, ACE-G with DINOv2 achieves the most balanced results across datasets. The accuracy of ACE and \methodname{} depends to some extent on the image features being used. The ACE features works well on static scenes that require little generalization but performs poorly in dynamic scenes. DINOv2 features are less precise compared to ACE features on static scenes, but generalize much better in dynamic scenes.

To further understand the differences on 7Scenes, we also include per-scene results on that dataset (\cref{tab:encoder}). 
The performance drop is caused by one scene (Stairs), and can be attributed to DINOv2 features, not to ACE-G's architecture or pre-training.

Notably, ACE-G's architecture and pre-training consistently improves when building on-top of DINOv2 features. The strong performance of ACE on static scenes, comes at the cost of worse performance on dynamic datasets. We believe that the strong performance of ACE-G in dynamic conditions is highly relevant in practice when query images are taken long after an environment has been mapped.

\subsection{Fine-Tuning}
In \cref{tab:datasetmix} we further show validation results for three specialized models fine-tuned on a subset of datasets for 1M iterations after 4M iterations of pre-training on all 6 datasets: in one case we fine-tune only on indoor datasets, in a second case only on outdoor datasets, and in the final case only on the MapFree dataset.
Interestingly, the latter models benefit less from the fine-tuning, which might suggest that the two outdoor training datasets (MapFree and BlendedMVG) are not sufficient for the variety of scenes present in these datasets.

\begin{table}[tb]
    \centering
    \caption{\textbf{Fine-tuning results.} Accuracy under $(20\si{\degree}, 20\,\mathrm{cm})$ error threshold on the validation splits of the training datasets. Fine-tuning on different dataset combinations can specialize the model for specific conditions.}%
    \vspace*{-0.5\baselineskip}
    \scriptsize%
    \renewcommand{\arraystretch}{1.1}%
    \setlength{\tabcolsep}{2pt}%
    \begin{tabular}{rC{0.7cm}C{0.7cm}C{0.7cm}C{0.7cm}C{0.7cm}C{0.7cm}}
        \toprule
        \quad ($\%$) & SN & SN++ & ARK & MF & BMVG & WR \\
        \midrule
        Baseline     & 63.5        & 55.5         & 56.0         & 47.1         & 40.6        & 36.7       \\
        Indoor       &\cellcolor{green!30}+1.9&\cellcolor{green!30}+2.6&\cellcolor{green!30}+1.2&\cellcolor{red!30}-3.4& \cellcolor{red!30}-7.5& \cellcolor{red!30}-9.2\\
        Outdoor      &\cellcolor{red!30}-8.7&\cellcolor{red!30}-10.6&\cellcolor{red!30}-9.6&\cellcolor{green!30}+0.4&\cellcolor{green!30}+3.4 &\cellcolor{red!30}-9.0\\
        MapFree &\cellcolor{red!30}-8.5 &\cellcolor{red!30}-10.1 &\cellcolor{red!30}{-9.1}&\cellcolor{green!30}{+0.2}&\cellcolor{red!30}{-7.1}&\cellcolor{red!30}{-8.6}\\        
        \bottomrule
    \end{tabular}
    \label{tab:datasetmix}
\end{table}

\end{document}